\documentclass[10pt,twocolumn,letterpaper]{article}

\usepackage[accsupp]{axessibility}
\usepackage{iccv}
\usepackage{times}
\usepackage{epsfig}
\usepackage{graphicx}
\usepackage{amsmath}
\usepackage{amssymb}

\usepackage{booktabs}
\usepackage{multirow}
\usepackage{color, colortbl}
\usepackage{subcaption}
\usepackage{tabu}
\usepackage{pifont}
\usepackage{makecell}
\usepackage{paralist}
\usepackage{threeparttable}
\usepackage{enumitem}
\usepackage{xcolor}

\def\Modelnamelight{UMT}
\def\Modelname{\textbf{UMT}}
\definecolor{Gray}{gray}{0.5}
\definecolor{LGray}{gray}{0.9}
\definecolor{darkblue}{RGB}{94,110,186}
\definecolor{darkGreen}{RGB}{92, 148, 110}
\definecolor{myblue}{RGB}{14, 121, 178}

\newcommand{\red}[1]{\textcolor{red}{#1}}
\newcommand{\blue}[1]{\textcolor{blue}{#1}}
\newcommand{\orange}[1]{\textcolor{orange}{#1}}

\newcommand{\gray}[1]{\textcolor{gray}{#1}}
\newcommand{\violet}[1]{\textcolor{violet}{#1}}
\newcommand{\darkGreen}[1]{\textcolor{darkGreen}{#1}}
\newcommand{\myblue}[1]{\textcolor{myblue}{#1}}
\newcommand{\darkblue}[1]{\textcolor{darkblue}{#1}}

\newcommand{\cmark}{\ding{51}}%
\newcommand{\xmark}{\ding{55}}%

\usepackage[pagebackref=true,breaklinks=true,letterpaper=true,colorlinks,bookmarks=false]{hyperref}

\iccvfinalcopy 

\ificcvfinal\fi

\begin{document}

\title{Unmasked Teacher: Towards Training-Efficient Video Foundation Models}

\author{
    Kunchang Li$^{1,2,3}$\thanks{Interns at Shanghai AI Laboratory. \textsuperscript{$\dag$}Corresponding authors.}\quad
    Yali Wang$^{1,3}$\textsuperscript{$\dag$}\quad
    Yizhuo Li$^{4,3}$\textsuperscript{*}\quad
    Yi Wang$^{3}$\quad
    Yinan He$^{3}$\\
    Limin Wang$^{5,3}$\quad
    Yu Qiao$^{3,1}$\textsuperscript{$\dag$}\vspace{0.2em}\\
    \small{$^1$Shenzhen Institute of Advanced Technology, Chinese Academy of Sciences}\\
    \small{$^2$University of Chinese Academy of Sciences\quad
    $^3$Shanghai AI Laboratory\quad
    $^4$The University of Hong Kong}\\
    \small{$^5$State Key Laboratory for Novel Software Technology, Nanjing University}\\
    \small{Code \& Models: \url{https://github.com/OpenGVLab/unmasked_teacher}}
}

\maketitle
\ificcvfinal\thispagestyle{empty}\fi

\begin{abstract}
Video Foundation Models (VFMs) have received limited exploration due to high computational costs and data scarcity. 
Previous VFMs rely on Image Foundation Models (IFMs), 
which face challenges in transferring to the video domain. 
Although VideoMAE has trained a robust ViT from limited data, 
its low-level reconstruction poses convergence difficulties and conflicts with high-level cross-modal alignment.
This paper proposes a training-efficient method for temporal-sensitive VFMs that integrates the benefits of existing methods.
To increase data efficiency,
we mask out most of the low-semantics video tokens,
but selectively align the unmasked tokens with IFM,
which serves as the \textbf{U}n\textbf{M}asked \textbf{T}eacher (\textbf{UMT}).
By providing semantic guidance,
our method enables faster convergence and multimodal friendliness.
With a progressive pre-training framework,
our model can handle various tasks including scene-related, temporal-related, and complex video-language understanding. 
Using only public sources for pre-training in \textbf{6 days} on \textbf{32 A100} GPUs,
our scratch-built ViT-L/16 achieves state-of-the-art performances on various video tasks.
\end{abstract}

\section{Introduction}

\begin{figure}[tp]
    \includegraphics[width=1.0\linewidth]{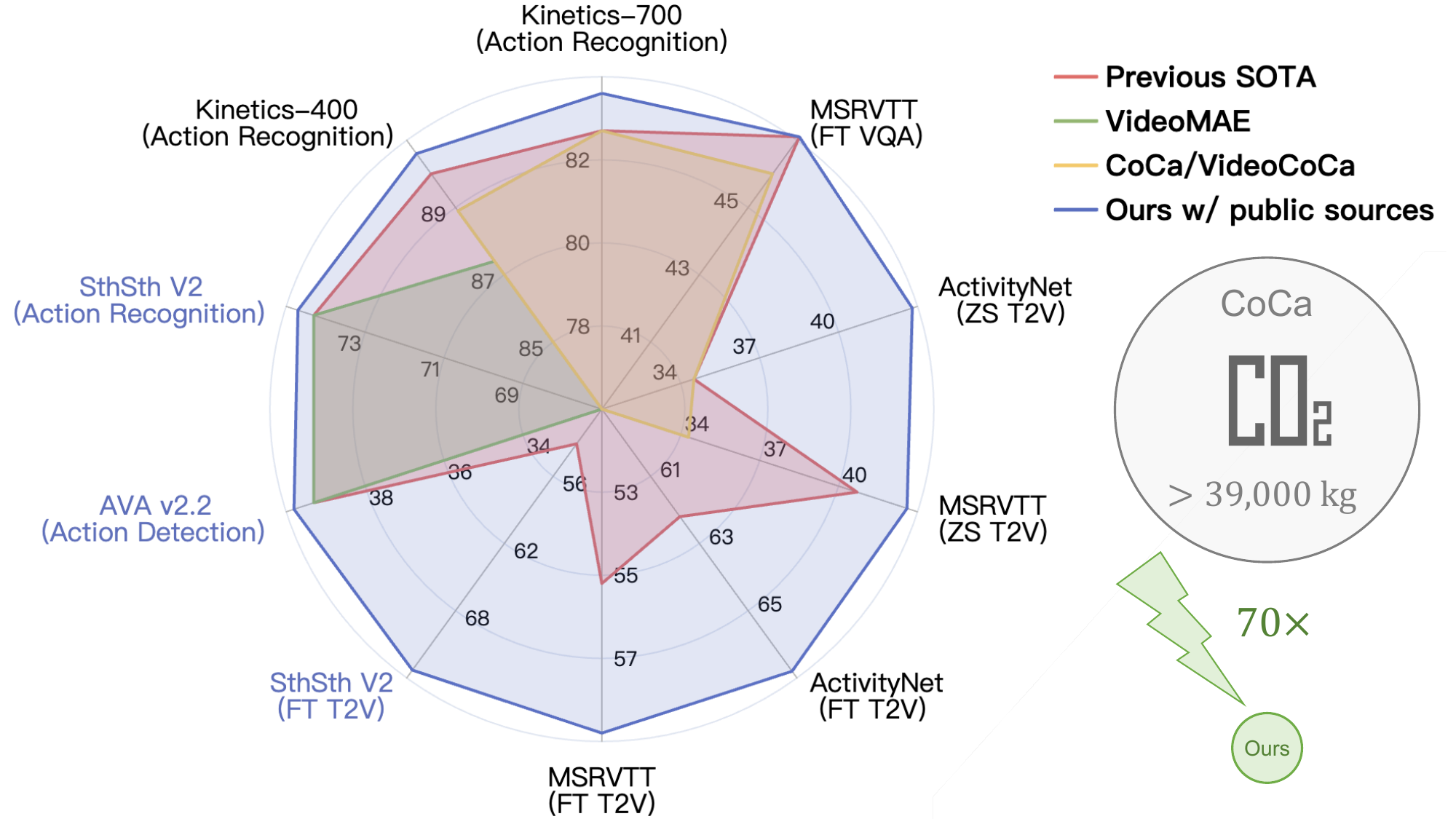}
    \vspace{-0.5cm}
    \caption{
    \textbf{Comparison with SOTA methods.}
    ``ZS'' and ``FT'' refer to ``zero-shot'' and ``fine-tuned''.
    ``T2V'' means video-text retrieval.
    For Kinetics action recognition,
    \cite{mtv} and \cite{Wang2022InternVideoGV} are excluded since they utilize model ensemble.
    With only public sources (\textit{i.e.}, CLIP\cite{clip}) for pre-training,
    our approach achieves SOTA performances on scene-related, \darkblue{temporal-related} and complex video-language benchmarks.
    Compared with CoCa \cite{coca},
    our method is much more environmentally friendly with \textbf{70$\times$} reduction in carbon emissions.
    Note that the cost of CLIP pre-training is ignored since it is publicly available.
    }
    \label{fig:intro}
    \vspace{-0.4cm}
\end{figure}

\begin{figure*}[thp]
    \centering
    \includegraphics[width=0.88\textwidth]{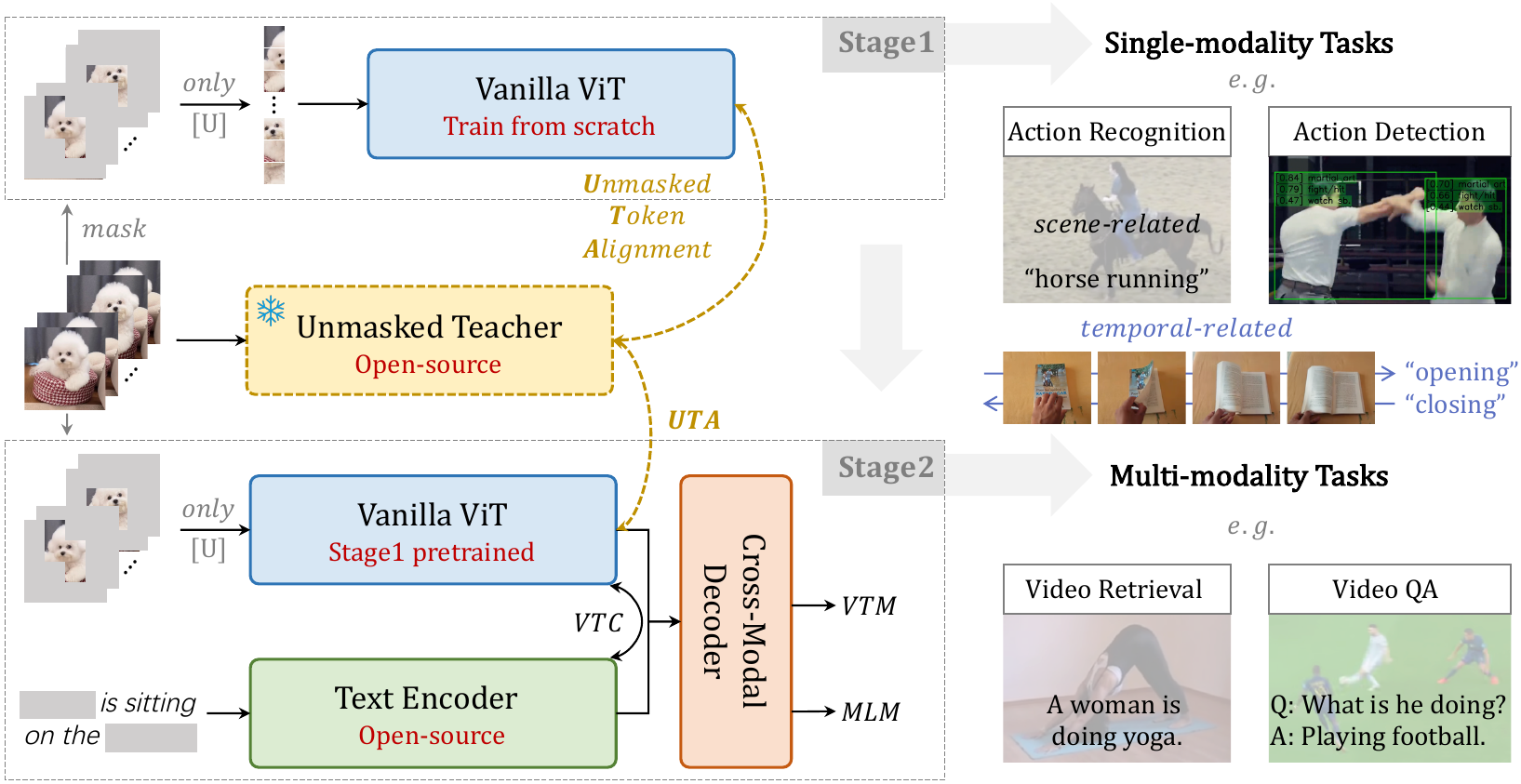}
    \vspace{-0.3cm}
    \caption{
    \textbf{Training-efficient framework for video foundation models.} 
    For general video understanding,
    we propose the \textit{progressive pre-training} with the unmasked teacher, which is \textit{simple, scalable and reproducible}. 
    The resulting models can not only handle scene-related and temporal-related actions well,
    but also conduct complex video-language understanding.
    }
    \label{fig:framework}
    \vspace{-0.3cm}
\end{figure*}

Video understanding has emerged as a critical skill for artificial intelligence systems to analyze and comprehend videos effectively. 
The progress in video understanding is currently driven by the Image Foundation Models (IFMs) \cite{vit,mae,beit,clip}, 
which are trained from massive datasets and adapted for different downstream tasks \cite{imagenet,ade,fickr,wang2023memoryandanticipation}. 
However, 
IFMs tend to focus more on scenes and objects, 
disregarding the essential motion patterns and object interactions required for complex video understanding. 
The \textit{true} Video Foundation Models (VFMs) are underexplored due to the high computational costs and data scarcity.

While building VFMs on well-learned IFMs reduces training costs, 
it poses significant challenges in transferring knowledge from the image domain to the video domain.
Firstly,
due to limited video data and a substantial domain gap,
video post-pretraining may undermine the generality inherited from IFMs \cite{xue2022clip}. 
Moreover, 
the strong spatial initialization offers a shortcut to perceive videos from scenes in single frames (\textit{e.g.}, ``grass'' in ``horse riding''),
which constrains VFMs from learning spatiotemporal relationships to recognize and localize \darkblue{temporal-related actions}, 
such as \darkblue{``opening" and ``closing"} in Figure \ref{fig:framework}.
Lastly, this paradigm is difficult to scale up as it requires well-prepared IFMs.

The recent success of VideoMAE \cite{videomae, st_mae} offers a data-efficient way to learn effective spatiotemporal features from scratch,
which handles complex temporal action recognition and detection tasks impressively. 
Nonetheless, 
its strong data efficiency and spatiotemporal modeling are traded by long pre-training (\textit{e.g.}, 2400 epochs on 160k videos). 
Besides, 
it is not well-suited for video-language tasks since the low-level pixel reconstruction task conflicts with high-level cross-modal alignment \cite{Shu2022MaskedCP}.
Additionally, 
the extra decoder that handles masked and unmasked tokens causes high memory costs due to global self-attention,
making scaling up this paradigm also challenging.

In this paper, 
we present a training-efficient method for temporal-sensitive VFMs by integrating the benefits of previous methods. 
Rather than directly adapting public IFM, \textit{e.g.}, CLIP \cite{clip}, 
we utilize them as \textbf{U}n\textbf{M}asked \textbf{T}eacher (\Modelname) to train vanilla ViT from scratch. 
We mask out most of the video tokens with low semantics and only align the unmasked tokens with a linear projection to the corresponding ones from the teacher. 
This approach not only inherits data efficiency from VideoMAE but also makes the learned video encoder multimodal-friendly (validated in Table \ref{tab:ablation_target}). 
Moreover,
training with only unmasked tokens without a decoder further saves GPU memory compared to VideoMAE,
and the guidance from the teacher's semantically rich representation leads to faster convergence. 
Notably, 
the resulting model can handle both scene-related \cite{k400,mit} and temporal-related actions \cite{sth,ava} exceptionally well, 
while the alignment to CLIP features enables the model to be compatible with cross-modal learning.

To address various video tasks,
we propose a progressive pre-training framework in Figure \ref{fig:framework}. 
In Stage 1, 
we only use video data for masked video modeling, 
resulting in a model that excels at video-only tasks.
In Stage 2, 
we employ public vision-language data for multi-modality learning. 
This allows the model to conduct complex video-language tasks, 
such as video-text retrieval \cite{msrvtt,lsmdc} and video question answering \cite{anet_qa,msrvtt_qa}. 
We use the \Modelnamelight\ in both stages, 
significantly reducing the training sources and speeding up convergence. 
Thanks to readily-available image and language foundation models \cite{clip,beitv2,lu2019vilbert,opt,flant5}, 
our simple framework is easily scalable for video foundation models.

We conduct extensive experiments to verify the effectiveness and efficiency of our approach.
As shown in Figure \ref{fig:intro},
with public sources (data/models) for pre-training,
our method achieves state-of-the-art performances on various video tasks,
including action recognition \cite{k400,k600,k700,mit,sth} (\textbf{90.6\%} top-1 accuracy on K400), 
spatiotemporal localization \cite{ava} (\textbf{39.8} mAP on AVA), 
video-text retrieval \cite{msrvtt,didemo,anet,lsmdc,msvd} (\textbf{58.8} R@1 on MSRVTT) and video question-answering \cite{anet_qa,msrvtt_qa,msrvtt_mc} (\textbf{47.1\%} accuracy on MSRVTT).
It is worth emphasizing that our method is much more environmentally friendly compared to CoCa \cite{coca}, which uses 2,048 CloudTPUv4 chips for 5 days. 
In contrast, 
our pre-training requires \textbf{32 A100(80G)} GPUs within \textbf{6 days}, 
leading to a remarkable \textbf{70$\times$} reduction in carbon emissions.
\section{Related Works}

\noindent\textbf{Video foundation models.}
The present Video Foundation Models (VFMs) are primarily based on well-prepared Image Foundation Models (IFMs) \cite{coca,mtv,vivit,cover,uniformerv2,fu2021violet,li2022lavender,wang2022omnivl,xue2022clip}. 
However, 
the strong spatial pre-training restricts their ability to learn spatiotemporal representations. 
Despite the impressive results demonstrated by Florence \cite{florence}, CoCa \cite{coca}, MTV \cite{mtv}, and UniFormerV2 \cite{uniformerv2} on video-only tasks \cite{k400,k600,k700}, 
these models struggle to handle temporal-related actions \cite{sth,finegym} and localize actions \cite{ava,thumos}. 
As for video-language tasks, 
there have been promising explorations on model architecture \cite{lei2021less,wang2022all,lei2022revealing} and learning paradigms \cite{xu2021videoclip,zellers2021merlot,fu2021violet,li2022lavender,wang2022omnivl}. 
Recently, 
InternVideo \cite{Wang2022InternVideoGV} introduces general VFMs through generative and discriminative learning.
However, 
the dependence on CLIP pre-training and tremendous training costs make it difficult to scale up.
In this paper, 
we propose an easily scalable framework for VFMs that is much more training-efficient.

\noindent\textbf{Masked vision modeling.} 
Inspired by the success of masked language modeling \cite{lu2019vilbert,dong2019unified}, 
masked vision modeling has been proposed for vision transformers \cite{vit}. 
BeiT \cite{bao2021beit} is the first to propose a BERT-like mask-then-predict framework to recover the discrete tokens \cite{ramesh2021zero}, 
while MAE \cite{mae} designs masked autoencoders to reconstruct normalized pixel values, 
which reduces memory consumption by processing only unmasked tokens in the encoder.
Later works can be roughly divided into BeiT-style \cite{peco,ibot,maskfeat,Baevski2022data2vecAG,maskdistill} and MAE-style \cite{simmim,Chen2022ContextAF,Gao2022ConvMAEMC,huang2022contrastive} with various target supervision, 
such as HOG descriptors \cite{maskfeat} and momentum features \cite{Tao2022SiameseIM}. 
For spatiotemporal learning, 
BEVT \cite{bevt} and VideoMAE \cite{videomae,st_mae} can be seen as extensions of BeiT and MAE, respectively. 
Recent works also indicate that CLIP features provide good guidance for mask modeling \cite{mvp,milan,maskdistill,beitv2,maskalign}, 
but all of them actually perform worse than CLIP itself with elaborate fine-tuning \cite{Dong2022CLIPII}. 
In contrast, 
we demonstrate that in the video domain,
our model with CLIP supervision clearly outperforms the teacher.
\section{Method}
In this section, 
we introduce our \textbf{U}n\textbf{M}asked \textbf{T}eacher (\Modelname) for masked video modeling and the progressive pre-training framework for temporal-sensitive video foundation models, 
as illustrated in Figure \ref{fig:framework}.

\subsection{Unmasked Teacher}
As discussed in the introduction, 
directly adapting the public Image Foundation Model (IFM) to Video Foundation Model (VFM) is challenging \cite{xclip,uniformerv2},
thus we propose using IFM as a teacher to train a VFM from scratch. 
Given the limited data scale, 
we leverage mask modeling \cite{mae} to make good use of the video data. 
However, 
unlike VideoMAE \cite{videomae}, 
we 
selectively align the unmasked tokens with the teacher,
removing an extra decoder for efficient training.

\textbf{Architecture.}
We choose CLIP-ViT \cite{clip} as an unmasked teacher due to its rich semantics that are learned with language guidance, 
which is beneficial for our following multi-modality learning. 
To fully impart the teacher's knowledge, 
we maintain its spatial architecture to process each video frame individually.
For our backbone, 
we apply the vanilla ViT without a class token. 
We employ spatiotemporal attention \cite{timesformer} to encourage all the unmasked tokens to interact with each other.
For better alignment with the spatial teacher, 
we do not use temporal downsampling,
thus the tokens can be aligned frame by frame.

\textbf{Masking.}
Following VideoMAE,
we use a high masking ratio (\textit{e.g.}, 80\%) to cut down video redundancies.
However,
the aggressive random masking may only retain the background tokens,
which contain insignificant information and hinder the teacher's knowledge transfer.
To enhance target effectiveness,
we apply the semantic masking \cite{milan} frame by frame,
where the tokens with important clues are maintained at higher probabilities.
Specifically,
given the class token $\mathbf{z}_{cls}$$\in$$\mathbb{R}^{1\times C}$ and the spatial tokens $\mathbf{Z}$$\in$$\mathbb{R}^{L\times C}$ in the $t$-th frame of CLIP-ViT ($L$$=$$H$$\times$$W$ is the token number and $C$ is the token dimension),
we calculate the attention score in the last self-attention \cite{vit} layer:
\begin{align}
\begin{split}
\mathbf{A} = {}& \sum_{n=1}^{N} \mathbf{A}_{n}(Q_{n}(\mathbf{\mathbf{z}_{cls}}), K_{n}(\mathbf{Z})) / N ,
\end{split}\\
\mathbf{A}_{n}(\mathbf{q}, \mathbf{k}) = {}& {\rm softmax}(\mathbf{q}\mathbf{k}^{T}/\sqrt{C/N}),
\end{align}
where $N$ is the head number,
and $Q_{n}(\cdot)$ and $K_{n}(\cdot)$ are the linear projections in the $n$-th head.
The $\mathbf{A}$$\in$$\mathbb{R}^{1\times L}$ represents the semantic importance of each token,
and we select the unmasked tokens by a multinomial distribution based on $\mathbf{A}$ to retain the informative objects in each frame.
Moreover,
we sparsely sample frames from the raw videos \cite{tsn},
which provides a more complicated action context due to the large frame stride.
The strategy encourages the model to reason long-term spatiotemporal relationships among objects.

\textbf{Target.}
For the teacher model, 
we input all $L$ spatial tokens along with the class token, 
frame by frame. 
In contrast, 
for the student model, 
we only input the unmasked tokens, 
which are equal to $L(1-r)T$ tokens, 
where $r$ is the masking ratio and $T$ is the frame number. 
To distill the rich semantics more effectively, 
we process the output teacher tokens using the pre-trained visual projection,
which is designed to establish meaningful connections between visual and text embeddings.
Additionally, 
we add a simple linear projection for the student model to align the token dimension. 
We select the corresponding unmasked token from the student and teacher, 
and compute the mean squared error (MSE) between the normalized pairs.
Compared to low-level pixel reconstruction,
token alignment requires a high-level understanding,
which is beneficial for multi-modality learning.

\subsection{Progressive Pre-training}
For general video understanding,
it is vital for the foundation model to handle video-language tasks.
However,
directly training such a model from scratch is inefficient.
For example,
CoCa \cite{coca} utilizes 4.8B data to train 5 days on 2,048 CloudTPUv4 chips.
Therefore,
we introduce a training-efficient framework with progressive pre-training.

\textbf{Pre-training pipeline.}
Figure \ref{fig:framework} outlines our pipeline.
In Stage 1, 
we train the ViT from scratch using only high-quality videos and guidance from Unmasked Teacher. 
The masked video modeling fully mines knowledge from the videos, 
resulting in a model that excels at video-only tasks.
In Stage 2, 
we equip the pre-trained ViT with a text encoder and cross-modal decoder,
initialized with the well-prepared language model. 
And we conduct multi-modality training with large-scale vision-text pairs, 
enabling the model to handle complex video-language tasks.
It's worth noting that currently, 
open-source language models are larger and more diverse than vision models,
making it easy to scale up our foundation models. 
For example, 
the largest OPT \cite{opt} has 175B parameters, 
while ViT-G \cite{Zhai2021ScalingVT} only has 1.8B.

\textbf{Pre-training objectives.}
For both stages, 
we utilize Unmasked Teacher to perform Unmasked Token Alignment (\textbf{UTA}). 
In Stage 2, 
we employ three other popular objectives:
\textbf{(i)} Video-Text Contrastive (\textbf{VTC}) learning, which aims to align the pooled unmasked video and text embeddings. 
We use the symmetric contrastive loss \cite{bain2021frozen} to maximize the mutual information.
\textbf{(ii)} Video-Text Matching (\textbf{VTM}) enhances cross-modal fusion by aligning the unmasked video and text tokens. 
We adopt the binary cross-entropy loss with hard negative mining \cite{li2021align,lei2022revealing}.
\textbf{(iii)} Masked Language Modeling (\textbf{MLM}) uses the cross-modal decoder to predict masked words from the other text and unmasked video tokens. 
We follow the BERT \cite{devlin2018bert} strategy but mask 50\% of the text tokens.
\section{Experiments}
\label{sec_exp}

\subsection{Implementation}

\textbf{Datasets.}
Unless otherwise stated,
we use Kinetics-710 dataset \cite{uniformerv2} in Stage 1, which is a combination of Kinetics-400, 600 and 700 \cite{k400,k600,k700} and excludes any repeated or leaked videos.
In Stage 2, 
we utilize image-text data for co-training \cite{wang2022all,lei2022revealing,wang2022omnivl},
where images are treated as single-frame videos. 
We use three corpora as in \cite{Cheng2022VindLUAR}:
\textbf{(i) 5M} Corpus comprises WebVid-2M \cite{bain2021frozen} video-text pairs and CC3M \cite{sharma2018conceptual} image-text pairs.
\textbf{(ii) 17M} Corpus includes four other image-text datasets: COCO \cite{lin2014microsoft}, Visual Genome \cite{krishna2017visual}, SBU Captions \cite{ordonez2011im2text}, and CC12M \cite{changpinyo2021conceptual}.
\textbf{(iii) 25M} Corpus uses a larger version of WebVid containing 10M video-text pairs.

\textbf{Settings.}
In this paper, 
we consider two model configurations: 
ViT-B/16 \cite{vit} with BERT$_{base}$ \cite{devlin2018bert} and ViT-L/16 with BERT$_{large}$. 
And CLIP-ViT-B/16 \cite{clip} and CLIP-ViT-L/14 are adopted as teachers for the base and large models, respectively.
Since CLIP-ViT-L/14 uses a smaller patch size,
we adopt a smaller input resolution (\textit{i.e.}, 196) to align the token number.
For Stage-1 pre-training, 
we follow most of the hyperparameter settings in VideoMAE \cite{videomae}. 
However, 
we sparsely sample \cite{tsn} 8 frames and use a masking ratio of 80\%.  
By default,
we train both models on 32 A100 with a batch size of 2048 for 200 epochs.
The training on Kinetics-710 takes about \textbf{60} and \textbf{90} hours for ViT-B/16 and ViT-L/16, respectively.
In Stage 2,
we follow \cite{lei2022revealing} to sample 4 frames and train for 10 epochs. 
Specifically,
we mask 50\% image and 80\% video tokens.
Both models are trained on 32 A100 with a batch size of 4096. 
The pre-training on 25M Corpus takes about \textbf{24} and \textbf{40} hours respectively for the base and large models. 
For more implementation details about training, 
please refer to the supplemental materials.

\begin{table}[tp]
    \centering
    \setlength\tabcolsep{4.0pt}
    \resizebox{0.9\linewidth}{!}{
        \begin{tabular}{ccc|c|cc|c}
        \textbf{[U]} & \textbf{[M]} & \textbf{MAE} & \textbf{Memory (G)} & \textbf{SSV2} & \textbf{K400} & \textbf{MSR}\\
        \Xhline{1.0pt}
        \xmark & \xmark & \cmark & 44.0 & 67.1 & 78.8 & 55.6 \\
        \cmark & \xmark & \cmark & 52.5 & \textbf{70.2} & 83.9 & 64.5 \\
        \cmark & \cmark & \xmark & 43.6 & 70.0 & 84.6 & 65.2 \\
        \rowcolor{gray!20} 
        \cmark & \xmark & \xmark & \textbf{16.0} & \textbf{70.2} & \textbf{84.9} & \textbf{66.8} \\
        \end{tabular}
    }
    \vspace{-0.3cm}
    \caption{\textbf{Target design.} 
    We benchmark ViT-B/16 in 32 A100 with a batch size of 2048.
    ``[U]'', ``[M]'' and ``MAE'' refers to unmasked token alignment, masked token recovering and pixel reconstruction~\cite{videomae} respectively.
    The pixel reconstruction conflict with our unmasked token alignment,
    and hinder the following multimodal learning.
    }
    \label{tab:ablation_target}
    \vspace{-0.3cm}
\end{table}
\begin{table}[tp]
    \centering
    \setlength\tabcolsep{4.5pt}
    \resizebox{0.8\linewidth}{!}{
        \begin{tabular}{ccc|cc}
        \textbf{Mask} & \textbf{Sampling} & \textbf{T-Down} & \textbf{SSV2} & \textbf{K400} \\
        \Xhline{1.0pt}
        Tube & Sparse & \xmark & \textbf{70.2} & 84.3 \\
        Random & Sparse & \xmark & \textbf{70.2} & 84.6 \\
        \rowcolor{gray!20} 
        Semantic & Sparse & \xmark & \textbf{70.2} & \textbf{84.9} \\
        Semantic & Dense & \xmark & 69.8 & 84.0 \\
        Semantic & Sparse & \cmark & 69.5 & 84.6 \\
        \end{tabular}
    }
    \vspace{-0.3cm}
    \caption{\textbf{Mask type, sampling method and temporal downsampling.} 
    Semantic masking~\cite{milan} works best.
    }
    \label{tab:ablation_mask_sampling_downsampling}
    \vspace{-0.3cm}
\end{table}

\subsection{Ablation Study}
We ablate the properties of \Modelnamelight\  in both stages on both scene-related \cite{k400,msrvtt} and temporal-related tasks \cite{sth,lei2022revealing}.
For single-modality learning, 
we pre-train ViT-B/16 for 200 epochs on SthSth V2 \cite{sth} or K400 \cite{k400} dataset. 
For multi-modality learning, 
we use K710 pre-trained models and further pre-train it for 10 epochs on 5M Corpus. 
Except for Table \ref{tab:ablation_target}, 
where we use K400 pre-training.

\textbf{Target.} 
Table \ref{tab:ablation_target} presents a comparison of training targets.
Compared with pixel reconstruction \cite{videomae},
our unmasked token alignment significantly improves the accuracy with only 36\% memory cost.
However,
combining the two targets results in poor results on K400 and MSRVTT,
indicating a conflict between low-level reconstruction and high-level alignment.
Moreover,
recovering the masked tokens has a detrimental effect,
possibly due to the high masking ratio making high-level recovery too challenging.
The results demonstrate our method is effective to learn temporal-sensitive and multimodal-friendly representation.

\textbf{Mask type, sampling method, and temporal downsampling.}
Table \ref{tab:ablation_mask_sampling_downsampling} indicates that different masking strategies yield comparable results in SthSth V2. 
We contend that recognizing the category of ``something" is not necessary for SthSth V2, but it requires deducing the intricate motion between objects, 
thus random masking suffices. 
However, 
it is critical for K400 to identify the scene and objects, 
making semantic masking advantageous for knowledge distillation. 
Moreover, 
sparse sampling without temporal downsampling is more appropriate for our approach.

\begin{figure}[tp]
    \centering
    \includegraphics[width=0.95\linewidth
    ]{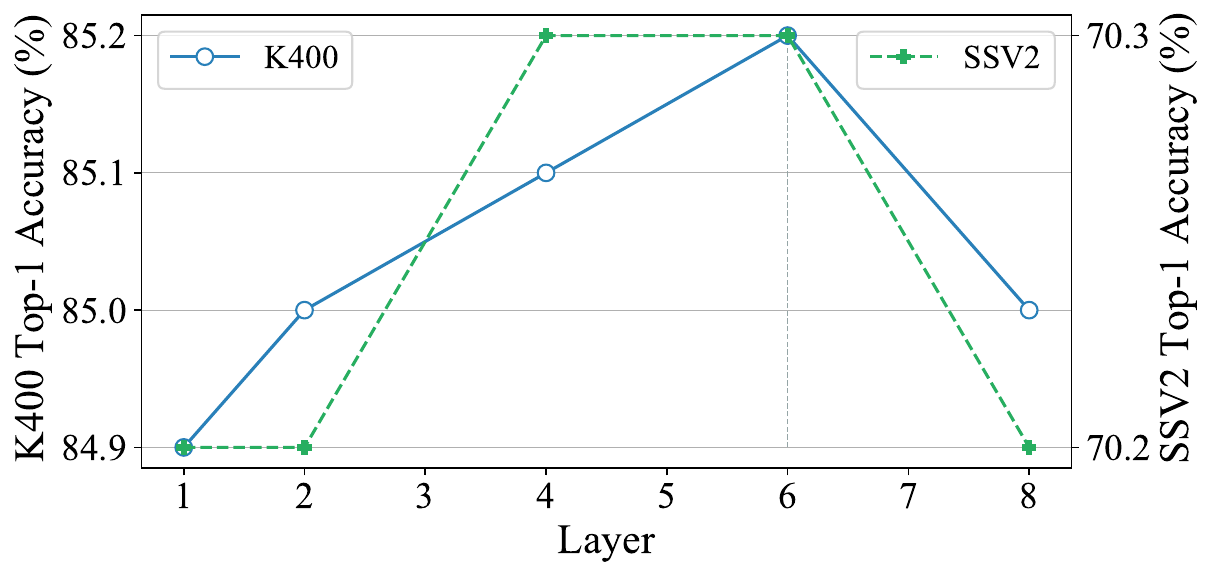}
    \vspace{-0.4cm}
    \caption{
    \textbf{Aligned layers.} 
    Since the GPU memory and running speed are similar,
    we align the last 6 layers.
    }
    \label{fig:ablation_layer}
    \vspace{-0.3cm}
\end{figure}

\textbf{Aligned layers.}
We try to align more layers in Figure \ref{fig:ablation_layer},
and the losses are averaged across multiple layers.
Since the GPU memory and running speed are similar,
we simply align the last 6 layers for the best results.

\begin{figure}[tp]
    \centering
    \includegraphics[width=0.95\linewidth
    ]{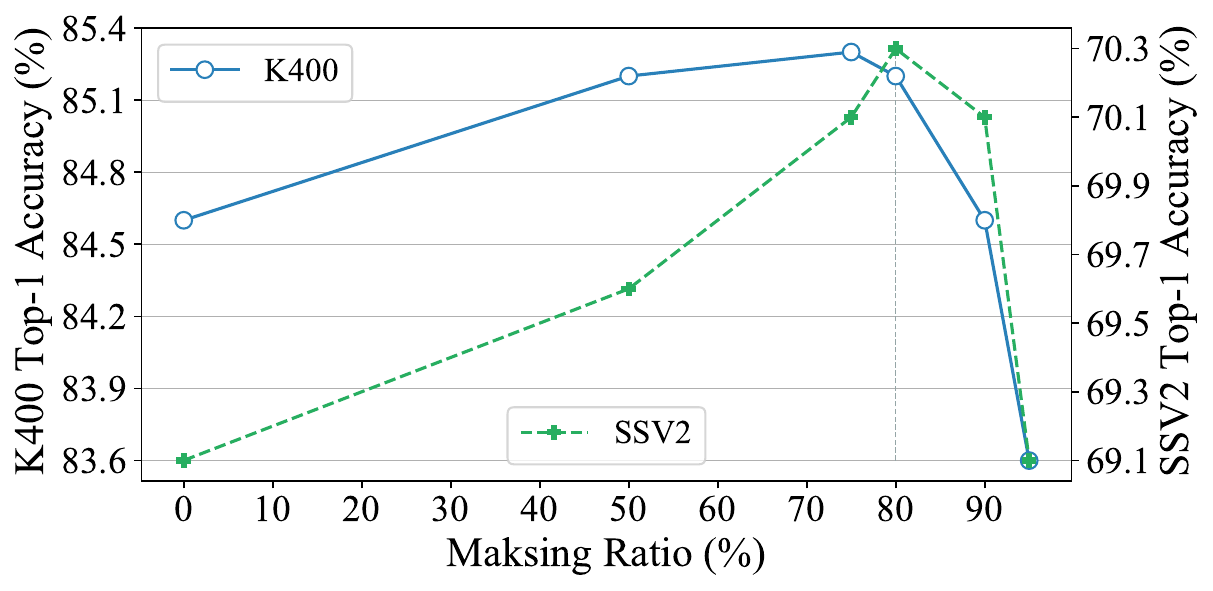}
    \vspace{-0.4cm}
    \caption{
    \textbf{Masking ratio.} 
    We use the masking ratio of 0.8 for a better trade-off on both datasets.
    }
    \label{fig:ablation_ratio}
    \vspace{-0.3cm}
\end{figure}

\textbf{Masking ratio.}
Figure \ref{fig:ablation_ratio} shows that proper high ratios work better.
When using a ratio of 95\%,
the performances dramatically drop since it is too challenging for token alignment.
Conversely,
when removing masks,
the task is too easy to learn the token relationships in space and time.
By default,
we adopt the ratio of 80\% for better trade-offs.


\begin{figure}[tp]
    \centering
    \begin{minipage}{\linewidth}
        \centering
        \includegraphics[width=0.83\linewidth]{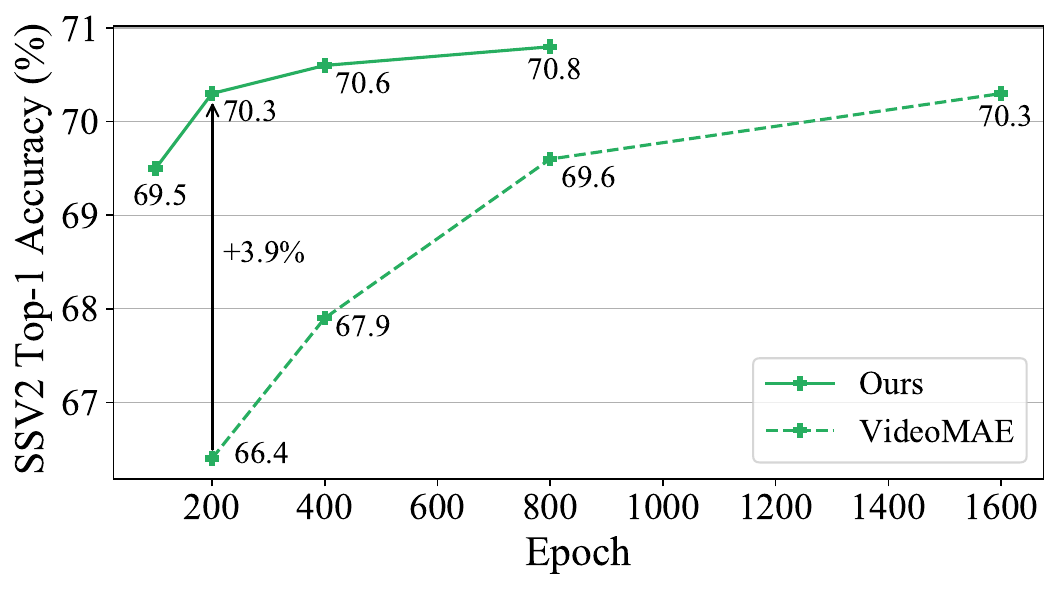}
        \vspace{-0.225cm}
        \subcaption{Results on SthSth V2}
    \end{minipage}
    \vspace{-0cm}
    \begin{minipage}{\linewidth}
        \centering
        \includegraphics[width=0.83\linewidth]{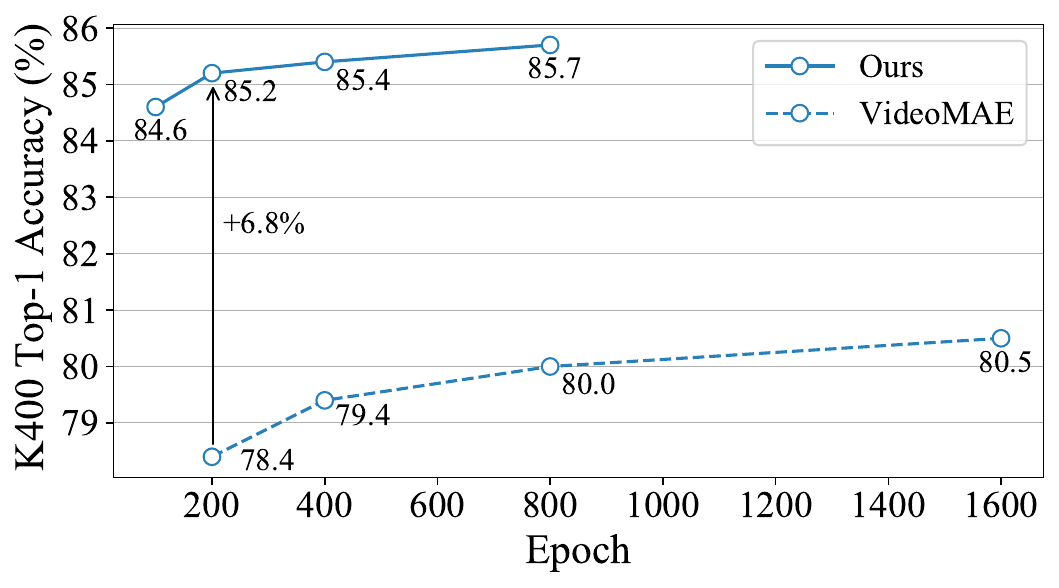}
        \vspace{-0.225cm}
        \subcaption{Results on Kinetics-400}
    \end{minipage}
\vspace{-0.3cm}
\caption{
\textbf{Training schedules.} 
A longer training schedule leads to more significant improvement.
}
\label{fig:ablation_epoch}
\vspace{-0.4cm}
\end{figure}

\textbf{Training schedule.}
Figure \ref{fig:ablation_epoch} presents the results of different training schedules.
On one hand,
a longer training schedule consistently improves the performances on both benchmarks.
On the other hand,
compared to VideoMAE \cite{videomae},
our method shows a faster convergence speed.
For example,
when pre-training for 200 epochs,
our models achieve 3.9\% and 6.8\% top-1 accuracy on SthSth V2 and Kinetics-400, respectively.

\begin{table}[tp]
    \centering
    \setlength\tabcolsep{4.0pt}
    \resizebox{1.0\linewidth}{!}{
        \begin{tabular}{l|c|l|l|cc}
        \textbf{Teacher} & \textbf{Mask} & \textbf{PT Student} & \textbf{FT Student} & \textbf{SSV2} & \textbf{K400} \\
        \Xhline{1.0pt}
        \rowcolor{green!3} 
        \multicolumn{4}{l|}{\textit{fine-tuning} CLIP-\textit{S}} & 57.4 & 82.0 \\
        \rowcolor{green!3} 
        \multicolumn{4}{l|}{\textit{fine-tuning} CLIP-\textit{ST}} & 68.0 & 82.5 \\
        \hline
        CLIP-\textit{S} & \xmark & ViT-\textit{S} & ViT-\textit{S} & 54.5 & 82.4 \\
        CLIP-\textit{S} & \cmark & ViT-\textit{S} & ViT-\textit{S} & 54.0 & 82.2 \\
        \hline
        CLIP-\textit{S} & \xmark & ViT-\textit{S} & ViT-\textit{ST} & 68.0 & 83.7 \\
        CLIP-\textit{S} & \cmark & ViT-\textit{S} & ViT-\textit{ST} & 67.2 & 83.4 \\
        \hline
        CLIP-\textit{S} & \xmark & ViT-\textit{ST} & ViT-\textit{ST} & 69.1 & 84.6 \\
        \rowcolor{gray!20} 
        CLIP-\textit{S} & \cmark & ViT-\textit{ST} & ViT-\textit{ST} & \textbf{70.3} & \textbf{85.2} \\
        \hline
        CLIP-\textit{ST} & \cmark & ViT-\textit{ST} & ViT-\textit{ST} & 68.7 & 83.7 \\
        \end{tabular}
    }
    \vspace{-0.3cm}
    \caption{\textbf{Why does \Modelnamelight\  work?} 
    ``\textit{S}'' and ``\textit{ST}'' refers to spatial and spatiotemporal attention respectively.
    Spatiotemporal attention and mask modeling are vital for \Modelnamelight. 
    }
    \label{tab:ablation_works}
    \vspace{-0.3cm}
\end{table}

\textbf{Why does \Modelnamelight\  work?}
In Table \ref{tab:ablation_works}, 
we investigate the crucial designs of our Unmasked Teacher. 
\textbf{(i) Spatiotemporal attention}: 
In the 2nd and 3rd parts, 
we compare the student with spatial attention and spatiotemporal attention during fine-tuning. 
Our results indicate that utilizing joint attention significantly enhances performance. 
Moreover, 
employing spatiotemporal attention during pre-training further improves performance (the 4th part), 
validating our assumption that joint attention encourages interaction among all unmasked tokens.
\textbf{(ii) Masked modeling}:
In the 4th part,
we observe that masked modeling plays a crucial role. 
However, 
when using spatial attention during pre-training, 
masked modeling becomes detrimental. 
We argue that when processing each frame individually with a high mask ratio of 80\%,
the token alignment task becomes excessively challenging.
\textbf{(iii) Teacher attention}:
The 5th part shows that although CLIP-\textit{ST} achieves better performance after fine-tuning, 
directly applying it as the teacher model leads to a performance drop. 
We contend that without post-training in the video domain, 
CLIP-\textit{ST} may disrupt the representation learned in the image domain.

\textbf{Outperforming the CLIP teacher.}
In the image domain,
the prior research \cite{Dong2022CLIPII} has shown that,
CLIP itself with fine-tuning surpasses existing CLIP-targeted MIM methods \cite{mvp,fd_clip,milan,maskdistill}.
However,
Table \ref{tab:ablation_works} indicates that in the video domain,
the student model (the 4th part) clearly outperforms the teacher,
\textit{i.e.},
CLIP-$ST$ with our elaborate fine-tuning.
We attribute the success to masked video modeling with spatiotemporal attention,
which encourages the model to capture long-term dependencies among objects.

\textbf{Different teachers.}
In Table \ref{tab:teacher},
we adopt different models \cite{dino,clip,beitv2} as the unmasked teachers.
As expected, 
the student models clearly outperform the corresponding teacher models, 
which have undergone elaborate fine-tuning. 
It's important to note that both student and teacher models share the same architecture, 
further emphasizing the effectiveness of our approach.

\begin{table}[tp]
    \centering
    \setlength\tabcolsep{7.5pt}
    \resizebox{0.9\linewidth}{!}{
        \begin{tabular}{l|l|ll}
        \textbf{Teacher} & \textbf{Student} & \multicolumn{1}{c|}{\textbf{SSV2}} & \multicolumn{1}{c}{\textbf{K400}} \\
        \Xhline{1.0pt}
        \rowcolor{green!3} 
        \multicolumn{2}{l|}{\textit{fine-tuning} DINO-\textit{ST}} & 65.0 & 80.8 \\
        DINO-\textit{ST} & ViT-\textit{ST} & 67.9 \red{(+2.9)} & 81.7 \red{(+0.9)} \\
        \hline
        \rowcolor{green!3} 
        \multicolumn{2}{l|}{\textit{fine-tuning} CLIP-\textit{ST}} & 68.0 & 82.5 \\
        CLIP-\textit{ST} & ViT-\textit{ST} & 70.3 \red{(+2.3)} & 85.2 \red{(+2.7)} \\
        \hline
        \rowcolor{green!3} 
        \multicolumn{2}{l|}{\textit{fine-tuning} BeiTv2-\textit{ST}} & 68.1 & 82.0 \\
        BeiTv2-\textit{ST} & ViT-\textit{ST} & 70.3 \red{(+2.2)} & 84.4 \red{(+2.4)} \\
        \end{tabular}
    }
    \vspace{-0.3cm}
    \caption{\textbf{Different teacher.}
    The student models significantly outperform the corresponding teacher models.
    }
    \label{tab:teacher}
    \vspace{-0.3cm}
\end{table}

\begin{table}[tp]
    \centering
    \setlength\tabcolsep{5.0pt}
    \resizebox{0.9\linewidth}{!}{
        \begin{tabular}{ccc|c|cc}
        \textbf{Image} & \textbf{Video} & \textbf{Text} & \textbf{Memory (G)} & \textbf{MSR} & \textbf{SSV2} \\
        \Xhline{1.0pt}
        50 & 60 & 50 & 35.6 & 66.9 & 80.6 \\
        \rowcolor{gray!20} 
        50 & 80 & 50 & 18.6 & \textbf{67.0} & \textbf{80.8} \\
        75 & 80 & 50 & 18.6 & 66.5 & 80.6 \\
        75 & 90 & 50 & 13.1 & 65.9 & 79.5 \\
        90 & 95 & 50 & 12.1 & 65.7 & 79.2 \\
        \hline
        50 & 80 & 25 & 18.6 & 66.5 & 80.1 \\
        50 & 80 & 75 & 18.6 & 66.6 & 78.2 \\
        \end{tabular}
    }
    \vspace{-0.3cm}
    \caption{\textbf{Different masking ratios for multi-modality pre-training.} 
    We benchmark ViT-B/16 in 16 A100 with a batch size of 2048.
    We report the average text-to-video retrieval R@1,5,10 accuracy of MSRVTT and SSV2-label.
    Masking 50\% image and 80\% video tokens works best.
    }
    \label{tab:ratio_mm}
    \vspace{-0.3cm}
\end{table}

\textbf{Multi-modality masking ratios.}
In Table \ref{tab:ratio_mm}, 
we first alter the masking ratios of the image and video data.
Since we co-train image-text and video-text data with the same batch size, 
the GPU memory primarily depends on the video masking ratio. 
As expected, 
processing images requires a lower masking ratio of 50\%. 
Although higher masking ratios reduce memory consumption, 
the corresponding performances are lower. 
Additionally, 
masking too few (25\%) or too many (75\%) text tokens leads to inferior results.

\begin{table}[tp]
    \centering
    \setlength\tabcolsep{4.5pt}
    \resizebox{1.0\linewidth}{!}{
        \begin{tabular}{cccc|c|cc}
        \textbf{VTC} & \textbf{VTM} & \textbf{MLM} & \textbf{UTA} & \textbf{Memory (G)} & \textbf{MSR} & \textbf{SSV2} \\
        \Xhline{1.0pt}
        \cmark & \xmark & \xmark & \cmark & 17.3 & 60.3 & 74.3 \\
        \xmark & \cmark & \xmark & \cmark & 18.1 & 65.6 & 79.1 \\
        \hline
        \cmark & \cmark & \xmark & \cmark & 18.2 & 65.9 & 79.2 \\
        \xmark & \cmark & \cmark & \cmark & 18.5 & 65.7 & 79.2 \\
        \rowcolor{gray!20} 
        \cmark & \cmark & \cmark & \cmark & 18.6 & \textbf{67.0} & \textbf{80.8} \\
        \hline
        \cmark & \cmark & \cmark & \xmark & 56.6 & 66.6 & 79.8 \\
        \end{tabular}
    }
    \vspace{-0.3cm}
    \caption{\textbf{Objectives for multi-modality pre-training.} 
    All the objectives are helpful to the downstream tasks.
    }
    \label{tab:objective_mm}
    \vspace{-0.3cm}
\end{table}

\begin{table}[tp]
    \centering
    \setlength\tabcolsep{6.0pt}
    \resizebox{0.9\linewidth}{!}{
        \begin{tabular}{l|cc}
        \textbf{Settings} & \textbf{MSR} & \textbf{SSV2} \\
        \Xhline{1.0pt}
        \rowcolor{gray!20} 
        Baseline & \textbf{67.0} & \textbf{80.8} \\
        one-stage pre-training & 57.9 & 64.5 \\
        only random mask w/o \Modelnamelight & 66.9 & 80.1 \\
        + visual projection alignment & 66.4 & 80.5 \\
        + visual \& text projection alignment & 66.0 & 80.0 \\
        + extra pre-training w/o mask & 66.5 & 80.5 \\
        \end{tabular}
    }
    \vspace{-0.3cm}
    \caption{\textbf{Other designs for multi-modality pre-training.} 
    }
    \label{tab:design_mm}
    \vspace{-0.3cm}
\end{table}

\textbf{Multi-modality pre-training objectives.}
For cross-modal retrieval,
utilizing either VTC or VTM for visual-text pairs is necessary.
In Table \ref{tab:objective_mm},
all loss weights are set to 1.
The 1st part reveals that VTM performs better than VTC.
Besides,
the 2nd part shows that combining VTC or MLM with VTM leads to a minor improvement,
while integrating all three objectives significantly enhances the performance.
Lastly,
without our unmasked teacher alignment,
the memory usage triples, 
while the performances drop.

\textbf{Other designs.}
Table \ref{tab:design_mm} showcases alternative designs for our multi-modality pre-training.
Firstly, 
we attempt to directly perform Stage 2 with a randomly initialized video encoder. 
For a fair comparison, 
we incorporate Kinetics-710 and conduct the same number of data iterations. 
However, 
the results demonstrate that the one-stage pre-training is challenging to converge, 
leading to poor performance. 
Secondly,
we randomly mask the video without an unmasked teacher for supervision,
which slightly reduces the overall performance. 
Additionally, 
we consider aligning the visual and text projection with the CLIP teacher,
since the teacher model also adopts contrastive learning. 
However, 
introducing extra alignment tasks turns out to be redundant and even harmful. 
Finally, 
we conduct extra pre-training without masks after masked pre-training.
Though it improves zero-shot performance (+1.5\% higher average recall accuracy),
the fine-tuned results are not as good as expected.

\begin{table*}[tp]
    \centering
    \setlength\tabcolsep{8.0pt}
    \resizebox{1.0\textwidth}{!}{
        \begin{tabular}{c|l|l|l|r|r|r|cc}
        \Xhline{1.0pt}
        \multicolumn{2}{l|}{\textbf{Method}} & \textbf{Backbone} & \textbf{Extra data} & \multicolumn{1}{c|}{\textbf{Input Size}} & \multicolumn{1}{c|}{\textbf{GFLOPs}}  & \multicolumn{1}{c|}{\textbf{Param}} & \textbf{Top-1}  & \textbf{Top-5} \\
        \Xhline{1.0pt}
        
        \multirow{14}{*}{\rotatebox{90}{\textit{supervised}}} & SlowFast$_{101}$~\cite{slowfast} &  R101+NL &  & 80$\times$224$^2$ & 234$\times$3$\times$10 & 60 & 79.8 & 93.9 \\
        ~ & MViTv2-B~\cite{mvitv2} & MViTv2-B &  & 32$\times$224$^2$ & 255$\times$1$\times$5 & 37 & 81.2 & 95.1 \\
        ~ & UniFormer-B\cite{uniformer} & UniFormer-B & IN-1K & 32$\times$224$^2$ & 259$\times$3$\times$4 & 50 & 83.0 & 95.4 \\
        ~ & TimeSformer-L~\cite{timesformer} & ViT-B & IN-21K & 96$\times$224$^2$ & 2380$\times$3$\times$1 & 121 & 80.7 & 94.7 \\
        ~ & VideoSwin-L~\cite{video_swin} & Swin-L &  IN-21K  & 32$\times$224$^2$ & 604$\times$3$\times$4 & 197 & 83.1 & 95.9  \\
        \cline{2-9}
        
        ~ & \multicolumn{8}{l}{\textit{Methods with web-scale data. FLD, ALIGN and CLIP consist of image-text pairs. WTS collects video-text pairs.}} \\
        ~ & ViViT-H~\cite{vivit}  & ViT-H & JFT-300M  & 32\blue{$\times$320$^2$} & 3981$\times$3$\times$4 & 654 & 84.9 & 95.8  \\
        ~ & CoVeR~\cite{cover}  & ViT-L & JFT-3B+SSV2+MiT+IN & 16\blue{$\times$448$^2$} & 5860$\times$3$\times$1 & 431 & 87.1 & -  \\
        ~ & CoCa~\cite{coca}  & ViT-g & JFT-3B+ALIGN-1.8B & 16\blue{$\times$576$^2$} & N/A$\times$3$\times$4 & 1000+ & 88.9 & -  \\
        ~ & MTV-H~\cite{mtv} & ViT-H+B+S+T & IN-21K+WTS-60M & 32\blue{$\times$280$^2$} & 6130$\times$3$\times$4 & 1000+ & 89.9 & 98.3 \\
        ~ & UniFormerV2-L~\cite{uniformerv2} & ViT-L & CLIP-400M+K710\red{$\dag$} & 64\blue{$\times$336$^2$} & 12550$\times$3$\times$2 & 354 & 90.0 & 98.4 \\
        \Xhline{0.8pt}

        \multirow{16}{*}{\rotatebox{90}{\textit{self-supervised}}} & BEVT$_{800e}$~\cite{bevt} & Swin-B & IN-1K & 32$\times$224$^2$ & 282$\times$3$\times$4 & 88 & 81.1 & - \\
        ~ & MaskFeat$_{1600e}$~\cite{maskfeat} & MViTv2-L &  & 16$\times$224$^2$ & 377$\times$1$\times$10 & 218 & 84.3 & 96.3 \\
        ~ & ST-MAE-B$_{1600e}$~\cite{st_mae} & ViT-B &  & 16$\times$224$^2$ & 180$\times$3$\times$7 & 87 & 81.3 & 94.9 \\
        ~ & ST-MAE-L$_{1600e}$~\cite{st_mae}  & ViT-L & K600 & 16$\times$224$^2$ & 598$\times$3$\times$7 & 304 & 84.9 & 96.2 \\
        ~ & ST-MAE-L$_{1600e}$~\cite{st_mae}  & ViT-L & K600\red{$\dag$} & 16$\times$224$^2$ & 598$\times$3$\times$7 & 304 & 86.5 & 97.2 \\
        ~ & VideoMAE-B$_{1600e}$~\cite{videomae} & ViT-B &  & 16$\times$224$^2$ & 180$\times$3$\times$5 & 87 & 81.5 & 95.1 \\
        ~ & VideoMAE-L$_{1600e}$~\cite{videomae} & ViT-L &  & 16$\times$224$^2$ & 597$\times$3$\times$5 & 305 & 85.2 & 96.8 \\
        ~ & VideoMAE-L$_{1600e}$~\cite{videomae} & ViT-L &  & 16$\times$320$^2$ & 3958$\times$3$\times$5 & 305 & 86.1 & 97.3 \\
        ~ & MVD-H$_{800e}$~\cite{mvd} & ViT-H & IN-1K & 16$\times$224$^2$ & 1192$\times$3$\times$5 & 633 & 87.2 & 97.4 \\
        \cline{2-9}
        
        ~ & \Modelname-B$_{800e}$ & ViT-B &  & 8$\times$224$^2$ & 180$\times$3$\times$4 & 87 & 85.7 & 97.0 \\
        ~ & \Modelname-B$_{200e}$ & ViT-B & K710 & 8$\times$224$^2$ & 180$\times$3$\times$4 & 87 & 85.7 & 96.9 \\
        ~ & \Modelname-B$_{200e}$ & ViT-B & K710\red{$\dag$} & 8$\times$224$^2$ & 180$\times$3$\times$4 & 87 & 87.4 & 97.5 \\
        ~ & \Modelname-L$_{400e}$ & ViT-L &  & 8$\times$224$^2$ & 596$\times$3$\times$4 & 304 & 88.9 & 98.3 \\
        ~ & \Modelname-L$_{200e}$ & ViT-L & K710 & 8$\times$224$^2$ & 596$\times$3$\times$4 & 304 & 89.1 & 98.2 \\
        ~ & \Modelname-L$_{200e}$ & ViT-L & K710\red{$\dag$} & 8$\times$224$^2$ & 596$\times$3$\times$4 & 304 & 90.3 & \textbf{98.7} \\
        ~ & \Modelname-L$_{200e}$ & ViT-L & K710\red{$\dag$} & 16$\times$224$^2$ & 1434$\times$3$\times$4 & 304 & \textbf{90.6} & \textbf{98.7} \\
        \Xhline{1.0pt}
        \end{tabular}
    }
    \vspace{-0.3cm}
    \caption{\textbf{Comparison with the state-of-the-art methods on Kinetics-400.} 
    For \Modelnamelight, we use a masking ratio of 80\%. 
    The results using spatial resolution $>$224$^2$ are noted in \blue{blue}.
     ``\red{$\dag$}'' marks the results with intermediate fine-tuning.}
    \label{tab:k400}
    \vspace{-0.6cm}
\end{table*}
\begin{table}[tp]
    \centering
    \setlength\tabcolsep{1.4pt}
    \resizebox{1.0\linewidth}{!}{
        \begin{tabular}{l|r|r|r|cc|cc}
        \Xhline{1.0pt}
        \multirow{2}{*}{\textbf{Method}} & \multicolumn{1}{c|}{\textbf{Input}} & \multicolumn{1}{c|}{\textbf{\small{FLOPs}}}  & \multicolumn{1}{c|}{\textbf{\small{Param}}} & \multicolumn{2}{c|}{\textbf{K600}} & \multicolumn{2}{c}{\textbf{K700}} \\
        ~ & \multicolumn{1}{c|}{\textbf{Size}} & \multicolumn{1}{c|}{\textbf{\small{(T)}}} & \multicolumn{1}{c|}{\textbf{\small{(M)}}} & \textbf{Top-1}  & \textbf{Top-5} & \textbf{Top-1}  & \textbf{Top-5} \\
        \Xhline{1.0pt}

        SlowFast$_{101}$~\cite{slowfast}  & 80$\times$224$^2$ & 7.0 & 60 & 81.8 & 95.1 & 71.0 & 89.6 \\
        MViTv2-L~\cite{mvitv2} & 40\blue{$\times$312$^2$} & 33.9 & 218 & 87.5 & 97.8 & 79.4 & 94.9\\
        \hline
        
        \multicolumn{8}{l}{\textit{Methods with web-scale data.}} \\
        CoVeR~\cite{cover} & 16\blue{$\times$448$^2$} & 17.6 & 431 & 87.9 & - & 79.8 & - \\
        CoCa~\cite{coca} & 16\blue{$\times$576$^2$} & N/A & 1000+ & 89.4 & - & 82.7 & - \\
        \scriptsize{UniFormerV2-L}~\cite{uniformerv2} & 32$\times$224$^2$ & 16.0 & 354 & 89.5 & 98.3 & 82.1 & 96.1 \\
        \scriptsize{UniFormerV2-L}~\cite{uniformerv2} & 64\blue{$\times$336$^2$} & 75.3 & 354 & 90.1 & 98.5 & 82.7 & 96.2 \\
        MTV-H~\cite{mtv} & 32$\times$224$^2$ & 44.5 & 1000+  & 89.6 & 98.3 & 82.2 & 95.7 \\
        MTV-H~\cite{mtv} & 32\blue{$\times$320$^2$} & 73.6 & 1000+  & 90.3 & 98.5 & 83.4 & 96.2 \\
        \Xhline{0.8pt}
        
        \Modelname-B & 8$\times$224$^2$ & 2.2 & 87 & 87.8 & 97.8 & 78.5 & 94.3 \\
        \Modelname-L & 8$\times$224$^2$ & 7.2 & 304 & 90.4 & 98.7 & 83.2 & 96.5 \\
        \Modelname-L & 16$\times$224$^2$ & 17.2 & 304 & \textbf{90.5} & \textbf{98.8} & \textbf{83.6} & \textbf{96.7} \\
        
        \Xhline{1.0pt}
        \end{tabular}
    }
    \vspace{-0.3cm}
    \caption{\textbf{Comparison with the state-of-the-art methods on Kinetics-600/700.}
    For \Modelnamelight, we report the results with K710 pre-training and intermediate fine-tuning.
    }
    \label{tab:k600_k700}
    \vspace{-0.4cm}
\end{table}
\begin{table}[tp]
    \centering
    \setlength\tabcolsep{3.53pt}
    \resizebox{1.0\linewidth}{!}{
        \begin{tabular}{l|r|r|r|cc}
        \Xhline{1.0pt}
        \multirow{2}{*}{\textbf{Method}} & \multicolumn{1}{c|}{\textbf{Input}} & \multicolumn{1}{c|}{\textbf{FLOPs}}  & \multicolumn{1}{c|}{\textbf{Param}} & \multicolumn{2}{c}{\textbf{MiT V1}} \\
        ~ & \multicolumn{1}{c|}{\textbf{Size}} & \multicolumn{1}{c|}{\textbf{(G)}} & \multicolumn{1}{c|}{\textbf{(M)}} & \textbf{Top-1}  & \textbf{Top-5} \\
        \Xhline{1.0pt}
        ViViT-L~\cite{vivit} & 32$\times$224$^2$ & 3980$\times$3 & 612 & 38.5 & 64.1 \\
        MTV-H~\cite{mtv} & 32$\times$224$^2$ & 3706$\times$12 & 1000+ & 45.6 & 74.7 \\
        \hline

        \multicolumn{6}{l}{\textit{Methods with web-scale data.}} \\
        CoVeR~\cite{cover} & 16\blue{$\times$448$^2$} & 5860$\times$3 & 431 & 46.1 & - \\
        MTV-H~\cite{mtv} & 32\blue{$\times$280$^2$} & 6130$\times$12 & 1000+ & 47.2 & 75.7 \\
        UniFormerV2-B~\cite{uniformerv2} & 8$\times$224$^2$ & 148$\times$12 & 115 & 42.7 & 71.5 \\
        UniFormerV2-L~\cite{uniformerv2} & 8$\times$224$^2$ & 666$\times$12 & 354 & 47.0 & 76.1 \\
        UniFormerV2-L~\cite{uniformerv2} & 8\blue{$\times$336$^2$} & 1568$\times$12 & 354 & 47.8 & 76.9 \\
        \Xhline{0.8pt}
        
        \Modelname-B & 8$\times$224$^2$ & 180$\times$12 & 87 & 44.6 & 74.0 \\
        \Modelname-L & 8$\times$224$^2$ & 596$\times$12 & 304 & 48.0 & 77.8 \\
        \hline
        \Modelname-B & 8\blue{$\times$384$^2$} & 786$\times$12 & 87 & 45.5 & 74.6 \\
        \Modelname-L & 8\blue{$\times$384$^2$} & 2440$\times$12 & 304 & \textbf{48.7} & \textbf{78.2} \\
        \Xhline{1.0pt}
        \end{tabular}
    }
    \vspace{-0.3cm}
    \caption{\textbf{Comparison with the state-of-the-art methods on Moments in Time V1.} 
    Following the previous methods,
    we fine-tune the models pre-trained by Kinetics-400.
    }
    \label{tab:mit}
    \vspace{-0.4cm}
\end{table}

\subsection{Single-modality tasks}
We evaluate our method on two conventional video-only tasks:
recognizing and localizing actions on six large-scale benchmarks,
including the \textit{Kinetics} family (\textit{i.e.}, Kinetics-400, 600 and 700 \cite{k400,k600,k700}),
\textit{Moments in Time} V1 \cite{mit} and \textit{Something-Something} V2 \cite{sth} for action recognition,
and \textit{AVA} V2.2 \cite{ava} for spatiotemporal localization.

\begin{table}[!t]
    \centering
    \setlength\tabcolsep{0.8pt}
    \resizebox{1.0\linewidth}{!}{
        \begin{tabular}{l|l|r|r|r|cc}
        \Xhline{1.0pt}
        \multirow{2}{*}{\textbf{Method}} & \multirow{2}{*}{\textbf{Extra Data}} & \multicolumn{1}{c|}{\multirow{2}{*}{\textbf{\#F}}} & \multicolumn{1}{c|}{\textbf{\small{FLOPs}}}  & \multicolumn{1}{c|}{\textbf{\small{Param}}} & \multicolumn{2}{c}{\textbf{SSV2}} \\
        ~ & ~ & ~ & \multicolumn{1}{c|}{\textbf{\small{(G)}}} & \multicolumn{1}{c|}{\textbf{\small{(M)}}} & \textbf{Top-1}  & \textbf{Top-5} \\
        \Xhline{1.0pt}
        \multicolumn{7}{l}{\textit{supervised}} \\
        SlowFast$_{101}$ \cite{slowfast} & K400 & 32 & 106$\times$3 & 53 & 63.1 & 87.6 \\
        TDN$_{EN}$~\cite{tdn} & IN-1K & 87 & 198$\times$3 & 88 & 69.6 & 92.2 \\
        \hline
        TimeSformer-L~\cite{timesformer} & IN-21K & 96 & 2380$\times$3 & 121 & 62.3 & - \\
        MViTv1-B~\cite{mvit} & K400 & 64 & 455$\times$3 & 37 & 67.7 & 70.9 \\
        MViTv2-B~\cite{mvitv2} & K400 & 64 &  225$\times$3 & 51 & 70.5 & 92.7 \\
        UniFormer-B~\cite{uniformer} & IN-1K+K400 & 32 & 259$\times$3 & 50 & 71.2 & 92.8 \\
        ViViT-L~\cite{vivit} & IN-21K+K400 & 32 & 3980$\times$3 & 612 & 65.9 & 89.9 \\
        MTV-B~\cite{mtv} & IN-21K+K400 & 32 & 399$\times$12 & 310 & 68.5 & 90.4 \\
        VideoSwin-B~\cite{video_swin} & IN-21K+K400  & 32 & 321$\times$3 & 88 & 69.6 & 92.7 \\
        \hline
        CoVeR \blue{$\uparrow$448}~\cite{cover} & JFT-3B+KMI & 16 & 5860$\times$3 & 431 & 69.9 & - \\
        UniFormerV2-B~\cite{uniformerv2} & CLIP-400M & 32 & 375$\times$3 & 163 & 70.7 & 93.2 \\
        UniFormerV2-L~\cite{uniformerv2} & CLIP-400M & 32 & 1718$\times$3 & 574 & 73.0 & 94.5 \\
        \Xhline{0.8pt}
        \multicolumn{7}{l}{\textit{self-supervised}} \\

        BEVT$_{800e}$~\cite{bevt} & IN-1K+K400 & 32 & 321$\times$3 & 88 & 70.6 & - \\
        \scriptsize{MaskFeat-L$_{1600e}$ \blue{$\uparrow$312}}~\cite{maskfeat} & K400 & 16 & 2828$\times$3 & 218 & 74.4 & 94.6 \\
        ST-MAE-L$_{1600e}$~\cite{st_mae} & K400\red{*} & 16 & 598$\times$3 & 304 & 72.1 & 93.9 \\
        ST-MAE-L$_{1600e}$~\cite{st_mae} & K600\red{*} & 16 & 598$\times$3 & 304 & 73.0 & 94.2 \\
        ST-MAE-L$_{1600e}$~\cite{st_mae} & K700\red{*} & 16 & 598$\times$3 & 304 & 73.6 & 94.4 \\
        VideoMAE-B$_{1600e}$~\cite{videomae} & K400\red{*} & 16 & 180$\times$6 & 87 & 69.7 & 92.3 \\
        VideoMAE-B$_{2400e}$~\cite{videomae} & - & 16 & 180$\times$6 & 87 & 70.8 & 92.4 \\
        VideoMAE-L$_{1600e}$~\cite{videomae} & K400\red{*}  & 16 & 597$\times$6 & 305 & 74.0 & 94.6 \\
        VideoMAE-L$_{2400e}$~\cite{videomae} & - & 16 & 597$\times$6 & 305 & 74.3 & 94.6 \\
        \hline
        \Modelname-B$_{800e}$ & - & 8 & 180$\times$6 & 87 & 70.8 & 92.6 \\
        \Modelname-B$_{200e}$ & K710\red{*} & 8 & 180$\times$6 & 87 & 70.8 & 92.4 \\
        \Modelname-L$_{400e}$ & - & 8 & 596$\times$6 & 304 & 74.4 & 94.5 \\
        \Modelname-L$_{200e}$ & K710\red{*} & 8 & 596$\times$6 & 304 & \textbf{74.7} & \textbf{94.7} \\
        \Xhline{1.0pt}
        \end{tabular}
    }
    \vspace{-0.3cm}
    \caption{\textbf{Comparison with the state-of-the-art methods on Something-Something V2.} ``\#F'' refers to the frame number. 
    ``KMI'' refers to ``K400+MiT+IN''. 
    ``\red{*}'' means the labels of extra data are not used for intermediate fine-tuning.
    }
    \label{tab:ssv2}
    \vspace{-0.4cm}
\end{table}

\textbf{Kinetics.}
Table \ref{tab:k400} reports the SOTA methods with supervised and self-supervised learning on K400.
On one hand,
our \Modelnamelight\  with intermediate fine-tuning outperforms the previous models that rely on web-scale pre-training,
\textit{e.g.},
the \Modelnamelight-L achieves 0.4\% higher top-1 accuracy than MTV-H \cite{mtv} with only 1/10 of the FLOPs and 1/3 of the parameters.
On the other hand,
our \Modelnamelight\  surpasses its counterparts with masked video modeling,
\textit{e.g.},
compared with VideoMAE \cite{videomae} with 1600-epoch pre-training,
the \Modelnamelight-L with 400-epoch pre-training obtains 3.7\% accuracy improvement.
For K600 and K700,
our \Modelnamelight-L also obtains the SOTA performances (\textbf{90.5\%} and \textbf{83.6\%} see Table \ref{tab:k600_k700}).

\textbf{Moments in Time.}
As shown in Table \ref{tab:mit},
our \Modelnamelight-L achieves \textbf{1.0\%/1.7\%} higher top-1/5 accuracy compared to the advanced UniFormerV2-L \cite{uniformerv2}, 
while utilizing fewer FLOPs.
Note that MiT is more challenging due to the large inter-class and intra-class variation, 
thus the results demonstrate the robustness and effectiveness of our method.

\textbf{Something-Something.}
Distinct from previous benchmarks,
this particular dataset requires complex and long-term modeling to accurately recognize temporal-related actions, 
such as "pretending to close something without actually closing it". 
Without any additional data, 
our \Modelnamelight-L model outperforms the UniFormerV2-L \cite{uniformerv2} (74.4\% \textit{vs.} 73.0\% in Table \ref{tab:ssv2}) which was specifically tailored for temporal modeling. 
Additionally, 
our approach achieves comparable performances to VideoMAE \cite{videomae} with significantly fewer epochs. 
Intriguingly, 
VideoMAE performs worse when utilizing Kinetics for masked modeling, 
while our \Modelnamelight\  performs even better. 
This demonstrates the versatility and adaptability of our method, 
which can be applied to diverse video domains with the same pre-training.

\textbf{AVA.}
Table \ref{tab:ava} presents the results of the action detection on AVA.
Remarkably, 
our \Modelnamelight\  achieves 2.0 mAP improvement over the advanced VideoMAE \cite{videomae} with only K400 pre-training.
Furthermore, 
our method achieves the impressive \textbf{39.8} mAP with K710 pre-training, 
showcasing its robust transferability for spatiotemporal understanding.

\begin{table}[!t]
    \centering
    \setlength\tabcolsep{4.0pt}
    \resizebox{1.0\linewidth}{!}{
        \begin{tabular}{l|l|r|r|r|c}
        \Xhline{1.0pt}
        \multirow{2}{*}{\textbf{Method}} & \multicolumn{1}{c|}{\textbf{PT}} & \multicolumn{1}{c|}{\textbf{Input}} & \multicolumn{1}{c|}{\textbf{FLOPs}}  & \multicolumn{1}{c|}{\textbf{Param}} & \multicolumn{1}{c}{\textbf{AVA}} \\
        ~ & \multicolumn{1}{c|}{\textbf{Data}} & \multicolumn{1}{c|}{\textbf{Size}} & \multicolumn{1}{c|}{\textbf{(G)}} & \multicolumn{1}{c|}{\textbf{(M)}} & \textbf{mAP} \\
        \Xhline{1.0pt}
        \multicolumn{6}{l}{\textit{supervised}} \\
        SlowFast~\cite{slowfast} & K400 & 32$\times$224$^2$ & 138 & 53 & 23.8 \\
        SlowFast~\cite{slowfast} & K600 & 64$\times$224$^2$ & 296 & 59 & 27.5 \\
        MViTv1-B~\cite{mvit} & K400 & 64$\times$224$^2$ & 455 & 36 & 27.3 \\
        MViTv1-B~\cite{mvit} & K600 & 32$\times$224$^2$ & 236 & 53 & 28.7 \\
        MViTv2-B~\cite{mvitv2} & K400 & 32$\times$224$^2$ & 225 & 51 & 28.1 \\
        MViTv2-B~\cite{mvitv2} & K700 & 32$\times$224$^2$ & 225 & 51 & 31.3 \\
        MViTv2-L~\cite{mvitv2} & \scriptsize{IN-21K+K700} & 40\blue{$\times$312$^2$} & 2828 & 213 & 33.5 \\
        \hline
        
        \multicolumn{6}{l}{\textit{self-supervised}} \\
        MaskFeat-L~\cite{maskfeat} & K400 & 40\blue{$\times$312$^2$} & 2828 & 218 & 36.3 \\
        MaskFeat-L~\cite{maskfeat} & K600 & 40\blue{$\times$312$^2$} & 2828 & 218 & 37.8 \\
        ST-MAE-L~\cite{st_mae} & K400 & 16$\times$224$^2$ & 598 & 304 & 34.8 \\
        ST-MAE-L~\cite{st_mae} & K700 & 16$\times$224$^2$ & 598 & 304 & 37.3 \\
        VideoMAE-B~\cite{videomae} & K400 & 16$\times$224$^2$ & 180 & 87 & 31.8 \\
        VideoMAE-L~\cite{videomae} & K400 & 16$\times$224$^2$ & 597 & 305 & 37.0 \\
        VideoMAE-L~\cite{videomae} & K700 & 16$\times$224$^2$ & 597 & 305 & 39.3 \\
        \Xhline{0.8pt}

        \Modelname-B & K400 & 8$\times$224$^2$ & 180 & 87 & 32.7 \\
        \Modelname-B & K710 & 8$\times$224$^2$ & 180 & 87 & 33.5 \\
        \Modelname-L & K400 & 8$\times$224$^2$ & 596 & 304 & 39.0 \\
        \Modelname-L & K710 & 8$\times$224$^2$ & 596 & 304 & \textbf{39.8} \\
        \Xhline{1.0pt}
        \end{tabular}

    }
    \vspace{-0.3cm}
    \caption{\textbf{Comparison with the state-of-the-art methods on AVA v2.2.} 
    All the self-supervised methods are with intermediate fine-tuning on the pre-training data.
    }
    \label{tab:ava}
    \vspace{-0.4cm}
\end{table}
\begin{table}[!t]
    \centering
    \small
    \setlength{\tabcolsep}{3pt}
    \resizebox{\linewidth}{!}{
    \begin{tabu}{lr|ccccc}
        \toprule
        {\bf Method} & {\bf \#Pairs} & \textbf{MSR} & \textbf{DDM} & \textbf{ANet} & \textbf{LSMDC} & \textbf{MSVD}  \\
        \midrule
        Frozen~\cite{bain2021frozen}  & 5M & 18.7 & 20.2 & - & - & -\\
        VIOLET~\cite{fu2021violet}  & 138M & 25.9 & 23.5 & - & - & - \\
        Singularity~\cite{lei2022revealing} & 17M & 34.0 & 37.1 & 30.6 & - & - \\
        OmniVL~\cite{wang2022omnivl} & 17M & 34.6 & 33.3 & - & - & -\\ 
        VINDLU~\cite{Cheng2022VindLUAR} & 25M & 32.0 & 36.9 & 30.9 & - & - \\
        \rowfont{\color{gray}}
        CLIP4Clip~\cite{luo2022clip4clip} & 400M & 30.6 & - & - & 13.6 & 36.2 \\
        \rowfont{\color{gray}}
        InternVideo~\cite{Wang2022InternVideoGV} & 646M & 40.7 & 31.5 & 30.7 & 17.6 & 43.4 \\ 
        \rowfont{\color{gray}}
        VideoCoCa~\cite{videococa} & 4.8B & 34.3 & - & 34.5 & - & - \\
        \midrule
        \multirow{3}{*}{\Modelname-B} & 5M & 29.6 & 33.4 & 28.3 & 16.8 & 36.2 \\
        & 17M & 35.5 & 41.9 & 33.8 & 18.1 & 41.4 \\
        & 25M & 35.2 & 41.2 & 35.5 & 19.1 & 42.3 \\
        \hline
        \multirow{3}{*}{\Modelname-L} & 5M & 33.3 & 34.0 & 31.9 & 20.0 & 44.4 \\
        & 17M & \textbf{42.6} & \underline{46.4} & \textbf{42.8} & \textbf{25.2} & \textbf{49.9} \\
        & 25M & \underline{40.7} & \textbf{48.6} & \underline{41.9} & \underline{24.9} & \underline{49.0} \\
        \bottomrule
    \end{tabu}
    }
    \vspace{-0.3cm}
    \caption{Zero-shot text-to-video retrieval on MSRVTT (``MSR''), DiDeMo (``DDM''), AcitivityNet (``ANet''), LSMDC, and MSVD.
    We only report the R@1 accuracy.
    Models pre-trained with large-scale pairs are noted in \gray{gray}.
    }
    \label{tab:retrieval_zs}
    \vspace{-0.3cm}
\end{table}

\begin{table*}
    \centering
    \small
    \setlength{\tabcolsep}{4.0
    pt}
    \resizebox{\textwidth}{!}{
    \begin{tabu}{lr|ccc|ccc|ccc|ccc|ccc}
        \toprule
        \multirow{2}{*}{\bf Method} & \multirow{2}{*}{\bf \#Pairs} & \multicolumn{3}{c|}{\bf MSRVTT} & \multicolumn{3}{c|}{\bf DiDeMo} & \multicolumn{3}{c|}{\bf ActivityNet} & \multicolumn{3}{c|}{\bf LSMDC} & \multicolumn{3}{c}{\bf MSVD}\\
        & & R@1 & R@5 & R@10 & R@1 & R@5 & R@10 & R@1 & R@5 & R@10 & R@1 & R@5 & R@10 & R@1 & R@5 & R@10\\
        \midrule
        ClipBERT~\cite{lei2021less}  & 5.4M & 22.0 & 46.8 & 59.9 & 20.4 & 48.0 & 60.8 & 21.3 & 49.0 & 63.5 & \multicolumn{3}{c|}{-} & \multicolumn{3}{c}{-} \\
        Frozen~\cite{bain2021frozen}  & 5M & 31.0 & 59.5 & 70.5 & 34.6 & 65.0 & 74.7 & \multicolumn{3}{c|}{-} & 15.0 & 30.8 & 39.8 & 33.7 & 64.7 & 76.3 \\
        VIOLET~\cite{fu2021violet}  & 138M & 34.5 & 63.0 & 73.4 & 32.6 & 62.8 & 74.7 & \multicolumn{3}{c|}{-} & 16.1 & 36.6 & 41.2 & \multicolumn{3}{c}{-} \\
        All-in-one~\cite{wang2022all} & 138M & 37.9 & 68.1 & 77.1 & 32.7 & 61.4 & 73.5 & 22.4 & 53.7 & 67.7 & \multicolumn{3}{c|}{-} & \multicolumn{3}{c}{-} \\
        LAVENDER~\cite{li2022lavender} & 30M & 40.7 & 66.9 & 77.6 & 53.4 & 78.6 & 85.3 & \multicolumn{3}{c|}{-} & 26.1 & 46.4 & 57.3 & 50.1 & 79.6 & 87.2\\
        Singularity~\cite{lei2022revealing} & 17M & 42.7 & 69.5 & 78.1 & 53.1 & 79.9 & 88.1 & 48.9 & 77.0 & 86.3 & \multicolumn{3}{c|}{-} & \multicolumn{3}{c}{-} \\
        OmniVL~\cite{wang2022omnivl} & 17M & 47.8 & 74.2 & 83.8 & 52.4 & 79.5 & 85.4 & \multicolumn{3}{c|}{-} & \multicolumn{3}{c|}{-} & \multicolumn{3}{c}{-}\\
        HiTeA~\cite{Ye2022HiTeAHT} & 17M & 46.8 & 71.2 & 81.0 & 56.5 & 81.7 & 89.7 & 49.7 & 77.1 & 86.7 & 28.7 & 50.3 & 59.0 & \multicolumn{3}{c}{-} \\ 
        VINDLU~\cite{Cheng2022VindLUAR} & 25M & 46.5 & 71.5 & 80.4 & 61.2 & 85.8 & 91.0 & 55.0 & 81.4 & 89.7 & \multicolumn{3}{c|}{-} & \multicolumn{3}{c}{-} \\
        \rowfont{\color{gray}}
        CLIP4Clip~\cite{luo2022clip4clip} & 400M & 44.5 & 71.4 & 81.6 & 42.8 & 68.5 & 79.2 & 40.5 & 72.4 & 83.4 & 21.6 & 41.8 & 49.8 & 46.2 & 76.1 & 84.6 \\
        \rowfont{\color{gray}}
        CLIP-ViP~\cite{xue2022clip} & 500M & 54.2 & 77.2 & 84.8 & 50.5 & 78.4 & 87.1 & 53.4 & 81.4 & 90.0 & 29.4 & 50.6 & 59.0 & \multicolumn{3}{c}{-} \\
        \rowfont{\color{gray}}
        InternVideo~\cite{Wang2022InternVideoGV} & 646M & 55.2 & 79.6 & 87.5 & 57.9 & 82.4 & 88.9 & 62.2  & 85.9 & 93.2 & 34.0 & 53.7 & 62.9 & 58.4 & 84.5 & 90.4 \\
        \midrule
        \multirow{3}{*}{\Modelname-B} & 5M & 46.3 & 72.7 & 82.0 & 54.8 & 83.0 & 89.0 & 52.1 & 80.5 & 89.6 & 30.3 & 51.8 & 61.4 & 47.4 & 76.8 & 84.0 \\
        & 17M & 50.6 & 75.4 & 83.5 & 60.8 & 85.1 & 91.0 & 56.1 & 82.5 & 91.2 & 32.3 & 54.5 & 61.9 & 49.6 & 78.5 & 85.7 \\
        & 25M & 51.0 & 76.5 & 84.2 & 61.6 & 86.8 & 91.5 & 58.3 & 83.9 & 91.5 & 32.7 & 54.7 & 63.4 & 50.8 & 79.7 & 86.2 \\
        \hline
        \multirow{3}{*}{\Modelname-L} & 5M & 53.3 & 76.6 & 83.9 & 59.7 & 84.9 & 90.8 & 58.1 & 85.5 & 92.9 & 37.7 & 60.6 & 67.3 & 53.7 & 80.5 & 86.8 \\
        & 17M & \underline{56.5} & \underline{80.1} & \textbf{87.4} & \underline{66.6} & \underline{89.9} & \textbf{93.7} & \underline{66.6} & \underline{88.6} & \underline{94.7} & \underline{41.4} & \underline{63.8} & \underline{72.3} & \underline{57.4} & \underline{83.0} & \underline{88.5} \\
        & 25M & \textbf{58.8} & \textbf{81.0} & \underline{87.1} & \textbf{70.4} & \textbf{90.1} & \underline{93.5} & \textbf{66.8} & \textbf{89.1} & \textbf{94.9} & \textbf{43.0} & \textbf{65.5} & \textbf{73.0} & \textbf{58.2} & \textbf{83.9} & \textbf{89.6} \\
        
        \bottomrule
    \end{tabu}
    }
    \vspace{-0.3cm}
    \caption{Text-to-video retrieval on MSRVTT, DiDeMo, AcitivityNet, LSMDC, and MSVD.
    ``\#Pairs'' denotes the number of pre-training pairs.
    Models pre-trained with large-scale pairs are noted in \gray{gray}.
    }
    \label{tab:retrieval_ft}
    \vspace{-0.5cm}
\end{table*}
\begin{table}[htpb]
    \centering
    \small
    \setlength{\tabcolsep}{3.0pt}
    \resizebox{1\linewidth}{!}{
    \begin{tabu}{lr|ccc|ccc}
        \toprule
        \multirow{2}{*}{\bf Method} & \multirow{2}{*}{\bf \#Pairs} & \multicolumn{3}{c|}{\bf SSV2-label} & \multicolumn{3}{c}{\bf SSV2-template} \\
        & & R@1 & R@5 & R@10 & R@1 & R@5 & R@10 \\
        \midrule
        \rowfont{\color{gray}}
        CLIP4Clip~\cite{luo2022clip4clip} & 400M & 43.1 & 71.4 & - & 77.0 & 96.6 & - \\
        Singularity~\cite{lei2022revealing} & 17M & 47.4 & 75.9 & - & 77.6 & 96.0 & - \\
        VINDLU~\cite{Cheng2022VindLUAR} & 25M & 53.1 & 81.8 & - & 83.3 & \textbf{100} & - \\
        HiTeA~\cite{Ye2022HiTeAHT} & 5M & 55.2 & 81.4 & 89.1 & 85.6 & \textbf{100} & \textbf{100} \\
        \midrule
        \multirow{3}{*}{\Modelname-B} & 5M & 63.1 & 87.1 & 92.3 & 87.3 & \textbf{100} & \textbf{100} \\
        & 17M & 63.4 & 88.0 & 92.9 & 86.8 & \underline{99.4} & \textbf{100} \\
        & 25M & 64.2 & 88.2 & 92.7 & 87.9 & \underline{99.4} & \textbf{100} \\
        \hline
        \multirow{3}{*}{\Modelname-L} & 5M & 70.5 & 92.3 & 95.5 & 90.2 & \underline{99.4} & \textbf{100} \\
        & 17M & 73.1 & \textbf{93.2} & \underline{96.4} & \textbf{90.8} & \textbf{100} & \textbf{100} \\
        & 25M & \textbf{73.3} & \underline{92.7} & \textbf{96.6} & \textbf{90.8} & \underline{99.4} & \textbf{100} \\
        \bottomrule
        
    \end{tabu}
    }
    \vspace{-0.3cm}
    \caption{Text-to-video retrieval on the temporally-heavy SSV2-label~\cite{lei2022revealing} and SSV2-template datasets~\cite{lei2022revealing}. 
    }
    \label{tab:retrieval_ssv2}
    \vspace{-0.3cm}
\end{table}

\subsection{Multi-modality tasks}
We further validate our model on two mainstream video-language tasks,
including video-text retrieval (MSRVTT \cite{msrvtt},
DiDeMo \cite{didemo}, ActivityNet \cite{anet}, LSMDC \cite{lsmdc}, MSVD \cite{msvd} and Something-Something \cite{lei2022revealing}) 
and video question-answering (ActivityNet-QA \cite{anet_qa}, MSRVTT-QA \cite{msrvtt_qa}, MSRVTT-MC \cite{msrvtt_mc} and MSVD-QA \cite{msrvtt_qa}).

\textbf{Zero-shot text-to-video retrieval.}
Table \ref{tab:retrieval_zs} indicates that the \Modelnamelight-B outperforms the top-performing models \cite{wang2022omnivl,lei2022revealing,Cheng2022VindLUAR} by \textbf{0.9\%}, \textbf{5.0\%}, and \textbf{4.6\%} R@1 on MSRVTT, DiDeMo, and ActivityNet, respectively.
In addition, our \Modelnamelight-L has set new records across all datasets. 
We see scores of \textbf{42.6\%}, \textbf{48.6\%}, \textbf{42.8\%}, \textbf{25.2\%}, and \textbf{72.2\%} on MSRVTT, DiDeMo, ActivityNet, LSMDC, and MSVD, in respective order. 
These notable results emphasize the exceptional robustness and effectiveness of our method.

\textbf{Text-to-video retrieval.}
Table \ref{tab:retrieval_ft} lists the fine-tuned results, 
where our \Modelnamelight-L significantly outperforms previous methods pre-trained with large-scale pairs \cite{clip4clip,xue2022clip,Wang2022InternVideoGV}. Specifically, 
our \Modelnamelight-L achieves \textbf{58.8\%} (+3.6\%), \textbf{70.4\%} (+9.2\%), \textbf{66.8\%} (+4.6\%), \textbf{43.0\%} (+9.0\%), and \textbf{80.3\%} (+21.9\%) on MSRVTT, DiDeMo, ActivityNet, LSMDC, and MSVD, respectively. 
Furthermore,
Table \ref{tab:retrieval_ssv2} showcases its impressive performances on the temporally-heavy SSV2-label and SSV2-template datasets, \textit{i.e.}, \textbf{73.3\%} and \textbf{90.8\%}, respectively.
These results demonstrate its profound capability for temporal modeling.

\textbf{Video question-answering.}
As shown in Table \ref{tab:vqa},
our \Modelnamelight\  outperforms the methods specifically designed for QA such as JustAsk \cite{yang2021just}, 
and achieves comparable performance with state-of-the-art models that pre-trained with large-scale pairs \cite{yang2022zero,Wang2022InternVideoGV,videococa},
which demonstrates its powerful capability of complex multimodal reasoning.

\begin{table}[thp]
    \centering
    \small
    \setlength{\tabcolsep}{2pt}
    \resizebox{1\linewidth}{!}{
    \begin{tabu}{l|r|cccc}
        \toprule
        \bf Method & \bf \#Pairs & \bf ANet & \bf MSR-QA & \bf MSR-MC & \bf MSVD-QA \\
        \midrule
        ClipBERT~\cite{lei2021less} & 0.2M & - & 37.4 & 88.2 & - \\
        ALPRO~\cite{li2022align} & 5M & - & 42.1 & - & 45.9 \\
        JustAsk~\cite{yang2021just} & 69M & 38.9 & 41.5 & - & 47.5 \\
        VideoCLIP~\cite{xu2021videoclip} & 136M & - & - & 92.1 & - \\
        All-in-one~\cite{wang2022all} & 138M & - & 44.3 & 92.0 & 47.9 \\
        MERLOT~\cite{zellers2021merlot} & 180M & 41.4 & 43.1 & 90.9 & - \\
        VIOLET~\cite{fu2021violet} & 138M & - & 43.9 & 91.9 & 47.9 \\
        Singularity~\cite{lei2022revealing} & 17M & 44.1 & 43.9 & 93.7 & - \\
        OmniVL~\cite{wang2022omnivl} & 17M & - & 44.1 & - & 51.0 \\
        VINDLU~\cite{Cheng2022VindLUAR} & 25M & 44.7 & 44.6 & 97.1 & - \\
        \rowfont{\color{gray}}
        FrozenBiLM~\cite{yang2022zero} & 400M & 43.2 & 47.0 & - & 54.8 \\
        \rowfont{\color{gray}}
        InternVideo~\cite{Wang2022InternVideoGV} & 646M & - & 47.1 & - & 55.5 \\
        \rowfont{\color{gray}}
        VideoCoCa~\cite{videococa} & 4.8B & - & 46.0 & - & 56.9 \\
        \midrule
        \multirow{3}{*}{\Modelname-B} & 5M & 43.5 & 44.3 & 95.9 & 49.1 \\
        & 17M & 44.9 & 44.9 & 96.3 & 48.9 \\
        & 25M & 44.8 & 44.9 & 96.3 & 49.5 \\
        \hline
        \multirow{3}{*}{\Modelname-L} & 5M & 45.1 & 45.5 & 96.8 & 51.3 \\
        & 17M & \underline{47.3} & \underline{46.4} & \textbf{97.7} & \underline{53.4} \\
        & 25M & \textbf{47.9} & \textbf{47.1} & \underline{97.3} & \textbf{55.2} \\
        \bottomrule
        
    \end{tabu}
    }
    \vspace{-0.3cm}
    \caption{Video question-answering on ActivityNet-QA, MSRVTT-QA, MSRVTT-MC and MSVD-QA. 
    }
    \label{tab:vqa}
    \vspace{-0.3cm}
\end{table}
\section{Conclusion}
In this paper, 
we propose using the image foundation model as the unmasked teacher for masked video modeling.
Besides,
we present a progressive pre-training framework for building environmentally friendly video foundation models,
which handles both scene-related and temporal-related actions, 
as well as complex video-language understanding. 
We hope that our simple, scalable, and reproducible framework will facilitate further research on video foundation models for future AI systems.
\section*{Acknowledgement}
This work was supported in part by the National Key R\&D Program of China (No. 2022ZD0160100, No. 2022ZD0160505, No. 2022ZD0160900), the National Natural Science Foundation of China (No. 62076119), the Joint Lab of CAS-HK, the National Natural Science Foundation of China under Grant (No. 62272450), the Shenzhen Research Program (RCJC20200714114557087), and in part by the Youth Innovation Promotion Association of Chinese Academy of Sciences (No. 2020355).

{\small
\bibliographystyle{ieee_fullname}
\bibliography{egbib}
}

\newpage
\appendix

\setcounter{page}{1}

\twocolumn[
\centering
\Large
\textbf{Unmasked Teacher: Towards Training-Efficient Video Foundation Models} \\
\vspace{0.5em}Supplementary Material \\
\vspace{1.0em}
] 

\begin{table}[tp]
    \centering
    \setlength\tabcolsep{8.0pt}
    \resizebox{1.0\linewidth}{!}{
        \begin{tabular}{c|c|c}
        \textbf{Stage} & \textbf{ViT-B} & \textbf{Output Size} \\
        \Xhline{1.0pt}
        Data & sparse sampling & \violet{3}$\times$\darkGreen{8}$\times$\myblue{224}$\times$\myblue{224} \\
        \hline
        Patch & \darkGreen{1}$\times$\myblue{16}$\times$\myblue{16}, \violet{768} & \multirow{2}{*}{\violet{768}$\times$\darkGreen{8}$\times$\orange{196}} \\
        Embedding & stride \darkGreen{1}$\times$\myblue{16}$\times$\myblue{16} & ~ \\
        \hline
        Position & sine-cosine & \multirow{2}{*}{\violet{768}$\times$\orange{1568}} \\
        Embedding & \violet{768}$\times$\orange{1568} & ~ \\
        \hline
        \multirow{2}{*}{Mask} & semantic mask & \multirow{2}{*}{\violet{768}$\times$\orange{1568$\cdot$(1-$\rho$)}} \\
        ~ & \textit{mask ratio} $=$ $\rho$ & ~ \\
        \hline
        Encoder & $\left[\begin{array}{c}\text{MHSA(\violet{768})}\\[-.1em] \text{MLP(\violet{3072})}\end{array}\right]$$\times$12 & \violet{768}$\times$\orange{1568$\cdot$(1-$\rho$)} \\
        \hline
        Projection & $\left[\begin{array}{c}\text{LN(\violet{768})}\\[-.1em] \text{MLP(\violet{512})}\end{array}\right]$$\times$$K$ & $K$$\times$\violet{512}$\times$\orange{1568$\cdot$(1-$\rho$)} \\
        \end{tabular}
    }
    \vspace{-0.3cm}
    \caption{\textbf{Architecture of video encoder.}
    We take ViT-B with 8-frame input as an example.
    ``MHSA'', ``MLP'' and ``LN'' refer to spatiotemporal multi-head self-attention, multi-layer perceptron and layer normalization.
    $K$ means the layer number for unmasked token alignment.
    We mark the \violet{channel number}, \darkGreen{frame number}, \myblue{spatial size} and \orange{token number} by different colors.
    }
    \label{tab:model_architecture}
\end{table}

\begin{table}[t!]
    \centering
    \setlength\tabcolsep{6pt}
    \resizebox{0.95\linewidth}{!}{
        \begin{tabular}{l|cc}
        config & SthSth V2 & Kinetics \\
        \Xhline{1.0pt}
        optimizer & \multicolumn{2}{c}{AdamW \cite{adamw}} \\ 
        optimizer momentum & \multicolumn{2}{c}{$\beta_1, \beta_2{=}0.9, 0.95$}  \\
        weight decay & \multicolumn{2}{c}{0.05} \\
        learning rate schedule & \multicolumn{2}{c}{cosine decay~\cite{cosine}} \\
        learning rate & \multicolumn{2}{c}{1.2e-3}\\
        batch size & \multicolumn{2}{c}{2048} \\
        warmup epochs \cite{warmup} & \multicolumn{2}{c}{40} \\
        total epochs &  \multicolumn{2}{c}{default 200} \\
        mask ratio & \multicolumn{2}{c}{default 80\%} \\
        input frame & \multicolumn{2}{c}{8} \\
        drop path \cite{droppath} & 0 & 0.1 (B), 0.2 (L) \\
        flip augmentation & \textit{no} & \textit{yes} \\
        augmentation & \multicolumn{2}{c}{MultiScaleCrop [0.66, 0.75, 0.875, 1]} \\
        \end{tabular}
    }
    \vspace{-0.3cm}
    \caption{
        \textbf{Stage-1 pre-training settings.}
    }
    \label{tab:stage1_hyperparameters} 
\end{table}

\section{More implementation details}

\subsection{Model architecture and training details}
In this section,
we introduce the model architectures and training hyperparameters in our experiments.

\textbf{Stage 1.}
In Stage 1,
we train the video encoder from scratch,
which is a vanilla ViT \cite{vit} without temporal downsampling.
We use the same patch size for both ViT-B and ViT-L,
\textit{i.e.}, $1$$\times$$16$$\times$$16$ ($T$$\times$$H$$\times$$W$).
To align with the unmasked teacher,
we use a simple linear projection,
including Layer Normalization \cite{ln} and one linear layer.
The example architecture is shown in Table \ref{tab:model_architecture}.
For pre-training,
we follow most of the hyperparameters in VideoMAE \cite{videomae},
as presented in Table \ref{tab:stage1_hyperparameters}.
However, 
to prevent overfitting, 
we use drop path \cite{droppath} in our approach.

\textbf{Stage 2.}
In Stage 2,
we equip the pre-trained video encoder with a text encoder and cross-modal decoder.
Following Singularity \cite{lei2022revealing},
for the base model,
we use the first 9 layers and the last 3 layers of BERT$_{base}$ to initialize the text encoder and decoder, respectively.
While for our large model,
we respectively adopt the first 19 layers and the 5 layers of BERT$_{large}$.
For pre-training,
we set all the loss weights to 1.
And more details are shown in Table \ref{tab:stage2_hyperparameters}.

\begin{table}[t!]
    \centering
    \setlength\tabcolsep{6pt}
    \resizebox{0.95\linewidth}{!}{
        \begin{tabular}{l|c}
        config & 5M \& 17M \& 25M \\
        \Xhline{1.0pt}
        optimizer & AdamW \cite{adamw} \\ 
        optimizer momentum & $\beta_1, \beta_2{=}0.9, 0.999$  \\
        weight decay & 0.02 \\
        learning rate schedule & cosine decay~\cite{cosine} \\
        learning rate & 1e-4\\
        batch size & 4096 (image), 4096 (video) \\
        warmup epochs \cite{warmup} & 1 \\
        total epochs &  10 \\
        mask ratio & 50\% (image), 80\% (video), 50\% (text) \\
        input frame & 4 \\
        drop path \cite{droppath} & 0.1 (B), 0.2 (L) \\
        flip augmentation & \textit{yes} \\
        augmentation & MultiScaleCrop [0.5, 1] \\
        \end{tabular}
    }
    \vspace{-0.3cm}
    \caption{
        \textbf{Stage-2 pre-training settings.}
    }
    \label{tab:stage2_hyperparameters} 
\end{table}
\begin{table}[t!]
    \centering
    \setlength\tabcolsep{2pt}
    \resizebox{1.0\linewidth}{!}{
        \begin{tabular}{l|ccc}
        config & SthSth & Kinetics & MiT \\
        \Xhline{1.0pt}
        optimizer & \multicolumn{3}{c}{AdamW \cite{adamw}} \\ 
        optimizer momentum & \multicolumn{3}{c}{$\beta_1, \beta_2{=}0.9, 0.999$}  \\
        weight decay & \multicolumn{3}{c}{0.05} \\
        learning rate schedule & \multicolumn{3}{c}{cosine decay~\cite{cosine}} \\
        learning rate & \small{4e-4 (B), 8e-4 (L)} & \small{4e-4 (B), 2e-4 (L)} & \small{1e-4 (B/L)} \\
        batch size & \multicolumn{3}{c}{512} \\
        repeated augmentation & 2 & 2 & 1 \\
        warmup epochs \cite{warmup} & 5 & 2 & 5 \\
        total epochs &  \small{30 (B), 17 (L)} & \small{35 (B), 20 (L)} & \small{40 (B), 20(L)} \\
        drop path \cite{droppath} & \multicolumn{3}{c}{0.1 (B), 0.2 (L)} \\
        layer-wise lr decay \cite{beit} & \multicolumn{3}{c}{0.75 (B), 0.85 (L)} \\
        flip augmentation & \textit{no} & \textit{yes} & \textit{yes} \\
        label smoothing \cite{label_smmoth} & \multicolumn{3}{c}{0.1} \\
        cutmix \cite{cutmix} & \multicolumn{3}{c}{1.0} \\
        augmentation & \multicolumn{3}{c}{RandAug(9, 0.5) \cite{randaugment}} \\
        \end{tabular}
    }
    \vspace{-0.3cm}
    \caption{
        \textbf{Action recognition fine-tuning settings.}
    }
    \label{tab:ar_hyperparameters} 
\end{table}

\textbf{Action Recognition.}
We adopt the Stage-1 pre-trained video encoder and add an extra classification layer for fine-tuning.
Detailed hyperparameters for different datasets are shown in Table \ref{tab:ar_hyperparameters}.
In our experiments,
we have tried to fine-tune the Stage-2 pre-trained video encoder,
but the results on Kinetics are similar.

\textbf{Action Detection.}
Following VideoMAE \cite{videomae} and ST-MAE \cite{st_mae},
we add ROIAlign with MaxPooling to generate the regions of interest.
Since we the Kinetics pre-trained models adopt sparse sampling \cite{tsn},
we use a frame span of 300 for action detection,
which is the default frame number of Kinetics videos.
More details are listed in Table \ref{tab:ad_hyperparameters}.

\textbf{Video-text retrieval.}
For fine-tuning,
we adopt the same architecture as in Stage 2,
but we only apply VTC and VTM losses.
For all datasets,
we sparsely sample 12 frames for both training and testing.
More details are listed in Table \ref{tab:ret_hyperparameters}.
For a fair comparison,
we follow Singularity \cite{lei2022revealing} to apply flip augmentation for SSV2 retrieval,
which may harm the performance of this temporal-related dataset.

\textbf{Video question-answering.}
Following the previous works \cite{lei2022revealing,Cheng2022VindLUAR,li2021align},
we formulate this task as text generation instead of classification.
We add an extra multi-modal decoder that takes the output of the cross-modal decoder as the keys/values.
And it decodes the answer text with ``[CLS]'' as a start.
We follow \cite{lei2022revealing,Cheng2022VindLUAR} to adopt the same architecture as the cross-modal decoder,
and initialize it using the pre-trained cross-modal decoder.
As for multiple-choice question-answering,
we follow \cite{lei2022revealing,li2021align,Cheng2022VindLUAR} to convert it to a text-to-video retrieval task,
where the question and candidate answers are concatenated.
The detailed hyperparameters are shown in Table \ref{tab:qa_hyperparameters} and Table \ref{tab:mc_hyperparameters}.

\begin{table}[t!]
    \centering
    \setlength\tabcolsep{8pt}
    \resizebox{0.7\linewidth}{!}{
        \begin{tabular}{l|c}
        config & AVA v2.2 \\
        \Xhline{1.0pt}
        optimizer & AdamW \cite{adamw} \\ 
        optimizer momentum & $\beta_1, \beta_2{=}0.9, 0.999$  \\
        weight decay & 0.05 \\
        learning rate schedule & cosine decay~\cite{cosine} \\
        learning rate & 1.25e-4 \\
        batch size & 128 \\
        warmup epochs \cite{warmup} & 5 \\
        total epochs &  30 (B), 25 (L) \\
        drop path \cite{droppath} & 0.2 (B), 0.4 (L) \\
        layer-wise lr decay \cite{beit} & 0.75 (B), 0.85 (L) \\
        flip augmentation & \textit{yes} \\
        \end{tabular}
    }
    \vspace{-0.3cm}
    \caption{
        \textbf{Action detection fine-tuning settings.}
    }
    \label{tab:ad_hyperparameters} 
\end{table}
\begin{table}[t!]
    \centering
    \setlength\tabcolsep{2pt}
    \resizebox{1.0\linewidth}{!}{
        \begin{tabular}{l|ccc}
        config & ActivityNet & MSRVTT & MSVD \\
        \Xhline{1.0pt}
        optimizer & \multicolumn{3}{c}{AdamW \cite{adamw}} \\ 
        optimizer momentum & \multicolumn{3}{c}{$\beta_1, \beta_2{=}0.9, 0.999$}  \\
        weight decay & \multicolumn{3}{c}{0.02} \\
        learning rate schedule & \multicolumn{3}{c}{cosine decay~\cite{cosine}} \\
        learning rate & 4e-5 (B/L) & 2e-5 (B/L) & 2e-5 (B) \\
        batch size & \multicolumn{3}{c}{256} \\
        warmup epochs \cite{warmup} & \multicolumn{3}{c}{1} \\
        total epochs &  12 (B), 10 (L) & 8 (B/L) & 15 (B), 6 (L) \\
        input frame & \multicolumn{3}{c}{12} \\
        drop path \cite{droppath} & 0.2 (B), 0.3 (L) & 0.2 (B), 0.4 (L) & 0.2 (B), 0.4 (L) \\
        flip augmentation & \multicolumn{3}{c}{\textit{yes}} \\
        augmentation & \multicolumn{3}{c}{MultiScaleCrop [0.5, 1]} \\
        \end{tabular}
    }
    \vspace{-0.3cm}
    \caption{
        \textbf{Video question-answering fine-tuning settings.}
    }
    \label{tab:qa_hyperparameters} 
\end{table}
\begin{table}[t!]
    \centering
    \setlength\tabcolsep{3.5pt}
    \resizebox{0.7\linewidth}{!}{
        \begin{tabular}{l|c}
        config & MSRVTT-MC \\
        \Xhline{1.0pt}
        optimizer & AdamW \cite{adamw} \\ 
        optimizer momentum & $\beta_1, \beta_2{=}0.9, 0.999$  \\
        weight decay & 0.02 \\
        learning rate schedule & cosine decay~\cite{cosine} \\
        learning rate & 8e-5 (B), 4e-5 (L)\\
        batch size & 256 \\
        warmup epochs \cite{warmup} & 0 \\
        total epochs &  5 \\
        input frame & 12 \\
        drop path \cite{droppath} & 0.2 (B), 0.3 (L) \\
        flip augmentation & \textit{yes} \\
        augmentation & MultiScaleCrop [0.5, 1] \\
        \end{tabular}
    }
    \vspace{-0.3cm}
    \caption{
        \textbf{Multi-choice video question-answering fine-tuning settings.}
    }
    \label{tab:mc_hyperparameters} 
\end{table}

\begin{table*}[t!]
    \centering
    \setlength\tabcolsep{4pt}
    \resizebox{1.0\linewidth}{!}{
        \begin{tabular}{l|ccccccc}
        config & MSRVTT & DiDeMo & ActivityNet & LSMDC & MSVD & SSV2-label & SSV2-template \\
        \Xhline{1.0pt}
        optimizer & \multicolumn{7}{c}{AdamW \cite{adamw}} \\ 
        optimizer momentum & \multicolumn{7}{c}{$\beta_1, \beta_2{=}0.9, 0.999$}  \\
        weight decay & \multicolumn{7}{c}{0.02} \\
        learning rate schedule & \multicolumn{7}{c}{cosine decay~\cite{cosine}} \\
        learning rate & 2e-5 (B/L) & 2e-5 (B), 4e-5 (L) & 4e-5 (B/L) & 2e-5 (B/L) & 2e-5 (B/L) & 5e-5 (B/L) & 1e-4 (B/L) \\
        batch size & \multicolumn{7}{c}{256} \\
        warmup epochs \cite{warmup} & \multicolumn{7}{c}{1} \\
        total epochs &  10 (B), 7(L) & 12 (B), 5 (L) & 20 (B/L) & 10 (B), 8 (L) & 10 (B/L) & 10 (B/L) & 10 (B), 8 (L)  \\
        input frame & \multicolumn{7}{c}{12} \\
        max text length & 32 & 64 & 150 & 96 & 64 & 25 & 25 \\
        drop path \cite{droppath} & 0.2 (B), 0.3 (L) & 0.1 (B), 0.3 (L) & 0.1 (B), 0.2 (L) & 0.1 (B), 0.2 (L) & 0.2 (B), 0.3 (L) & 0.1 (B), 0.2 (L) & 0.1 (B), 0.2 (L) \\
        flip augmentation & \multicolumn{7}{c}{\textit{yes}} \\
        augmentation & \multicolumn{7}{c}{MultiScaleCrop [0.5, 1]} \\
        \end{tabular}
    }
    \vspace{-0.3cm}
    \caption{
        \textbf{Video-text retrieval fine-tuning settings.}
    }
    \label{tab:ret_hyperparameters} 
\end{table*}
\begin{table*}
    \centering
    \small
    \setlength{\tabcolsep}{4.0
    pt}
    \resizebox{\textwidth}{!}{
    \begin{tabu}{lr|c|ccc|ccc|ccc|ccc|ccc}
        \Xhline{1.0pt}
        \multirow{2}{*}{\bf Method} & \multirow{2}{*}{\bf \#Pairs} & \multirow{2}{*}{\bf Type} & \multicolumn{3}{c|}{\bf MSRVTT} & \multicolumn{3}{c|}{\bf DiDeMo} & \multicolumn{3}{c|}{\bf ActivityNet} & \multicolumn{3}{c|}{\bf LSMDC} & \multicolumn{3}{c}{\bf MSVD}\\
        ~ & ~ & ~ &  R@1 & R@5 & R@10 & R@1 & R@5 & R@10 & R@1 & R@5 & R@10 & R@1 & R@5 & R@10 & R@1 & R@5 & R@10\\
        \Xhline{0.8pt}
        \multirow{6}{*}{\Modelname-B} & \multirow{2}{*}{5M} & T2V & 29.6 & 52.8 & 61.9 & 33.4 & 58.3 & 67.0 & 28.3 & 53.0 & 64.2 & 16.8 & 30.5 & 37.6 & 36.2 & 65.7 & 76.1 \\
        ~ & ~ & V2T & 26.2 & 46.7 & 54.9 & 32.0 & 58.7 & 68.2 & 25.9 & 50.2 & 61.7 & 12.9 & 27.4 & 33.6 & 58.5 & 78.7 & 84.3 \\
        \cline{3-18}
        ~ & \multirow{2}{*}{17M} & T2V & 35.5 & 59.3 & 68.6 & 41.9 & 66.7 & 75.0 & 33.8 & 59.1 & 70.4 & 18.1 & 33.1 & 40.0 & 41.4 & 70.6 & 80.1 \\
        ~ & ~ & V2T & 31.6 & 53.5 & 64.1 & 40.3 & 66.6 & 75.8 & 31.6 & 56.2 & 67.9 & 16.0 & 29.9 & 35.7 & 62.5 & 80.8 & 87.0 \\
        \cline{3-18}
        ~ & \multirow{2}{*}{25M} & T2V & 35.2 & 57.8 & 66.0 & 41.2 & 65.4 & 74.9 & 35.5 & 60.6 & 71.8 & 19.1 & 33.4 & 42.2 & 42.3 & 71.7 & 80.8 \\
        ~ & ~ & V2T & 30.3 & 50.7 & 61.4 & 40.8 & 67.7 & 76.7 & 32.8 & 57.6 & 69.2 & 15.7 & 30.6 & 37.4 & 61.9 & 82.5 & 88.5 \\
        \hline
        \multirow{6}{*}{\Modelname-L} & \multirow{2}{*}{5M} & T2V & 33.3 & 58.1 & 66.7 & 34.0 & 60.4 & 68.7 & 31.9 & 60.2 & 72.0 & 20.0 & 37.2 & 43.7 & 44.4 & 73.3 & 82.4 \\
        ~ & ~ & V2T & 30.2 & 51.3 & 61.6 & 36.2 & 60.0 & 68.6 & 30.0 & 59.1 & 71.3 & 16.1 & 32.0 & 39.2 & 66.1 & 85.5 & 89.4 \\
        \cline{3-18}
        ~ & \multirow{2}{*}{17M} & T2V & \textbf{42.6} & \textbf{64.4} & \textbf{73.1} & 46.4 & 70.0 & 78.8 & \textbf{42.8} & \textbf{69.6} & 79.8 & \textbf{25.2} & \textbf{43.0} & 50.5 & \textbf{49.9} & \textbf{77.7} & \textbf{85.3} \\
        ~ & ~ & V2T & \textbf{38.6} & \textbf{59.8} & \textbf{69.6} & 46.5 & 72.2 & 79.5 & \textbf{40.7} & \textbf{67.6} & \textbf{78.6} & \textbf{23.2} & 37.7 & 44.2 & \textbf{75.4} & \textbf{89.6} & \textbf{94.0} \\
        \cline{3-18}
        ~ & \multirow{2}{*}{25M} & T2V & 40.7 & 63.4 & 71.8 & \textbf{48.6} & \textbf{72.9} & \textbf{79.0} & 41.9 & 68.9 & \textbf{80.3} & 24.9 & 41.7 & \textbf{51.8} & 49.0 & 76.9 & 84.7\\
        ~ & ~ & V2T & 37.1 & 58.7 & 68.9 & \textbf{49.9} & \textbf{74.8} & \textbf{81.4 }& 39.4 & 66.8 & 78.3 & 21.9 & \textbf{37.8} & \textbf{45.7} & 74.5 & 89.7 & 92.8 \\
        \Xhline{1.0pt}
    \end{tabu}
    }
    \vspace{-0.3cm}
    \caption{\textbf{Zero-shot} retrieval results on MSRVTT, DiDeMo, AcitivityNet, LSMDC, and MSVD.
    }
    \label{tab:more_retrieval_zs}
\end{table*}
\begin{table*}
    \centering
    \small
    \setlength{\tabcolsep}{4.0
    pt}
    \resizebox{\textwidth}{!}{
    \begin{tabu}{lr|c|ccc|ccc|ccc|ccc|ccc}
        \Xhline{1.0pt}
        \multirow{2}{*}{\bf Method} & \multirow{2}{*}{\bf \#Pairs} & \multirow{2}{*}{\bf Type} & \multicolumn{3}{c|}{\bf MSRVTT} & \multicolumn{3}{c|}{\bf DiDeMo} & \multicolumn{3}{c|}{\bf ActivityNet} & \multicolumn{3}{c|}{\bf LSMDC} & \multicolumn{3}{c}{\bf MSVD}\\
        ~ & ~ & ~ &  R@1 & R@5 & R@10 & R@1 & R@5 & R@10 & R@1 & R@5 & R@10 & R@1 & R@5 & R@10 & R@1 & R@5 & R@10\\
        \Xhline{0.8pt}
        \multirow{6}{*}{\Modelname-B} & \multirow{2}{*}{5M} & T2V & 46.3 & 72.7 & 82.0 & 54.8 & 83.0 & 89.0 & 52.1 & 80.5 & 89.6 & 30.3 & 51.8 & 61.4 & 47.4 & 76.8 & 84.0 \\
        ~ & ~ & V2T & 44.4 & 72.8 & 80.7 & 52.9 & 80.2 & 85.8 & 50.0 & 79.8 & 88.2 & 29.8 & 52.2 & 60.5 & 69.1 & 85.8 & 92.1 \\
        \cline{3-18}
        ~ & \multirow{2}{*}{17M} & T2V & 50.6 & 75.4 & 83.5 & 60.8 & 85.1 & 91.0 & 56.1 & 82.5 & 91.2 & 32.3 & 54.5 & 61.9 & 49.6 & 78.5 & 85.7 \\
        ~ & ~ & V2T & 49.4 & 76.7 & 83.5 & 59.5 & 83.8 & 90.7 & 54.6 & 82.1 & 91.1 & 31.5 & 53.6 & 61.9 & 71.6 & 88.8 & 92.7 \\
        \cline{3-18}
        ~ & \multirow{2}{*}{25M} & T2V & 51.0 & 76.5 & 84.2 & 61.6 & 86.8 & 91.5 & 58.3 & 83.9 & 91.5 & 32.7 & 54.7 & 63.4 & 50.8 & 79.7 & 86.2 \\
        ~ & ~ & V2T & 49.0 & 77.0 & 84.7 & 59.5 & 84.9 & 90.5 & 56.0 & 83.5 & 91.7 & 32.7 & 53.5 & 63.2 & 73.3 & 89.6 & 93.7 \\
        \hline
        \multirow{6}{*}{\Modelname-L} & \multirow{2}{*}{5M} & T2V & 53.3 & 76.6 & 83.9 & 59.7 & 84.9 & 90.8 & 58.1 & 85.5 & 92.9 & 37.7 & 60.6 & 67.3 & 53.7 & 80.5 & 86.8 \\
        ~ & ~ & V2T & 51.4 & 76.3 & 82.8 & 59.5 & 84.5 & 90.7 & 55.4 & 84.4 & 92.9  & 36.2 & 58.9 & 65.7 & 77.2 & 91.6 & 94.8 \\
        \cline{3-18}
        ~ & \multirow{2}{*}{17M} & T2V & 56.5 & 80.1 & 87.4 & 66.6 & 89.9 & 93.7 & 66.6 & 88.6 & 94.7 & 41.4 & 63.8 & 72.3 & 57.4 & 83.0 & 88.5 \\
        ~ & ~ & V2T & 56.7 & 79.6 & 86.7 & \textbf{66.4} & 87.5 & 92.9 & 64.3 & 87.8 & \textbf{94.8} & 40.3 & 63.1 & 71.1 & 82.4 & 93.6 & 96.0 \\
        \cline{3-18}
        ~ & \multirow{2}{*}{25M} & T2V & \textbf{58.8} & \textbf{81.0} & \textbf{87.1} & \textbf{70.4} & \textbf{90.1} & \textbf{93.5} & \textbf{66.8} & \textbf{89.1} & \textbf{94.9} & \textbf{43.0} & \textbf{65.5} & \textbf{73.0} & \textbf{58.2} & \textbf{83.9} & \textbf{89.6} \\
        ~ & ~ & V2T & \textbf{58.6} & \textbf{81.6} & \textbf{86.5} & 65.7 & \textbf{89.6} & \textbf{93.3} & \textbf{64.4} & \textbf{89.1} & \textbf{94.8} & \textbf{41.4} & \textbf{64.3} & \textbf{71.5} & \textbf{82.4} & \textbf{94.6} & \textbf{96.7} \\
        \Xhline{1.0pt}
    \end{tabu}
    }
    \vspace{-0.3cm}
    \caption{\textbf{Fine-tuned} retrieval results on MSRVTT, DiDeMo, AcitivityNet, LSMDC, and MSVD.
    }
    \label{tab:more_retrieval_ft}
    \vspace{-0.3cm}
\end{table*}

\section{More results}

\subsection{Video-text retrieval}
Table \ref{tab:more_retrieval_zs} and Table \ref{tab:more_retrieval_ft} show more zero-shot and fine-tuned retrieval results on MARVTT \cite{msrvtt}, DiDeMo \cite{didemo}, ActivityNet \cite{anet}, LSMDC \cite{lsmdc} and MSVD \cite{msvd}.

\subsection{Dataset descriptions}
We show the statistics of pre-training datasets in Table \ref{tab:statics_pretrain},
and downstream datasets in Table \ref{tab:statics_downstream}.

\begin{table*}[tp]
    \centering
    \setlength\tabcolsep{8.0pt}
    \resizebox{0.7\textwidth}{!}{
        \begin{tabular}{lccc}
        \toprule
        \textbf{Dataset} & \textbf{\#image/video} & \textbf{\#text} & \textbf{Type} \\
        \midrule
        Kinetics-710 \cite{uniformerv2} & 658K & 0 & Video \\
        COCO \cite{lin2014microsoft} & 113K & 567K & image \\
        Visual Genome \cite{krishna2017visual} & 100K & 768K & image \\
        SBU Captions \cite{ordonez2011im2text} & 860K & 860K & image \\
        CC3M \cite{sharma2018conceptual} & 2.88M & 2.88M & image \\
        CC12M \cite{changpinyo2021conceptual} & 11.00M & 11.00M & image \\
        WebVid-2M \cite{bain2021frozen} & 2.49M & 2.49M & video \\
        WebVid-10M \cite{bain2021frozen} & 10.73M & 10.73M & video \\
        \midrule
        5M corpus = CC3M$+$WebVid-2M & 5.37M & 5.37M & video+image \\
        17M corpus = 5M$+$COCO$+$VG$+$SBU$+$CC12M & 17.44M & 18.57M & video+image \\
        25M corpus = 17M$+$WebVid-10M$-$WebVid-2M & 25.68M & 26.81M & video+image \\
        \bottomrule
        \end{tabular}
    }
    \vspace{-0.3cm}
    \caption{\textbf{Statistics of pre-training datasets.}
    }
    \label{tab:statics_pretrain}
    \vspace{-0.1cm}
\end{table*}

\begin{table*}[tp]
    \centering
    \setlength\tabcolsep{15.0pt}
    \resizebox{1.0\textwidth}{!}{
        \begin{tabular}{lrrrrrrr}
        \toprule
        \multirow{2}{*}{ \textbf{Dataset} } & \multicolumn{3}{c}{\textbf{\#video}} & \multicolumn{3}{c}{\textbf{\#text}} & \textbf{Avg Video} \\  \cmidrule{2-4} \cmidrule(l){5-7}
        & \textbf{Train} & \textbf{Val} & \textbf{Test} & \textbf{Train} & \textbf{Val} & \textbf{Test} & \textbf{Length (s)} \\ 
        \midrule
        \multicolumn{3}{l}{\textit{Action Recognition}} & & & & & \\
        Kinetics-400 \cite{k400} & 240,436 & 19,787 & - & - & - & - & 10 \\
        Kinetics-600 \cite{k600} & 366,006 & 27,935 & - & - & - & - & 10 \\
        Kinetics-700 \cite{k700} & 529,573 & 33,861 & - & - & - & - & 10 \\
        Moments in Time V1 \cite{mit} & 802,244 & 33,899 & - & - & - & - & 3 \\
        Something-Something V2 \cite{sth} & 168,913 & 24,777 & - & - & - & - & 4 \\
        \midrule
        \multicolumn{3}{l}{\textit{Action Detection}} & & & & & \\
        AVA v2.2 \cite{ava} & 235 & 64 & 131 & - & - & - & 900 \\
        \midrule
        \multicolumn{3}{l}{\textit{Video-Text Retrieval}} & & & & & \\
        MSRVTT \cite{msrvtt} & 7,010 & - & 1,000 & 140,200 & - & 1,000 & 15 \\
        DiDeMo \cite{didemo} & 8,496 & 1,094 & 1,036 & 8,496 & 1,094 & 1,036 & 29.3 \\
        ActivityNet Captions~\cite{anet} & 10,009 & 4,917 & - & 10,009 & 4,917 & - & 180 \\
        LSMDC \cite{lsmdc} & 101,055 & - & 1,000 & 101,055 & - & 1,000 & 4.7 \\
        MSVD \cite{msvd} & 1,200 & 100 & 670 & 1,200 & 100 & 670 & 15 \\
        SSV2-Template \cite{lei2022revealing} & 168,913 & - & 2,088 & 174 & - & 174 & 4 \\
        SSV2-Label \cite{lei2022revealing} & 168,913 & - & 2,088 & 109,968 & - & 1,989 & 4 \\
        \midrule
        \multicolumn{3}{l}{\textit{Video Question-Answering}} & & & & & \\
        ActivityNet-QA \cite{anet_qa} & 3,200 & 1,800 & 800 & 32,000 & 18,000 & 8,000 & 180 \\
        MSRVTT-QA \cite{msrvtt_qa} & 6,513 & 497 & 2,990 & 158,581 & 12,278 & 72,821 & 15 \\
        MSRVTT-MC \cite{msrvtt_mc} & 7,010 & - & 2,990 & 140,200 & - & 14,950 & 15 \\
        MSVD-QA \cite{msrvtt_qa} & 1,161 & 245 & 504 & 29,883 & 6,415 & 13,157 & 15  \\
        \bottomrule
        \end{tabular}
    }
    \vspace{-0.3cm}
    \caption{\textbf{Statistics of downstream datasets.}
    }
    \label{tab:statics_downstream}
    \vspace{-0.1cm}
\end{table*}

\end{document}